 \documentclass[final,5p,times,twocolumn]{elsarticle}

\usepackage{amssymb}
\usepackage{lipsum}
\usepackage{graphicx}
\usepackage[dvipsnames]{xcolor}
\usepackage{CJKutf8}
\usepackage{xcolor}
\usepackage{subcaption}
\usepackage{float}
\usepackage{booktabs}
\usepackage{array}
\usepackage{hyperref}

\newcommand{\boxedref}[1]{\fcolorbox{red}{white}{\ref{#1}}}

\journal{Nuclear Physics B}

\begin{document}
\begin{CJK*}{UTF8}{gbsn}
\begin{frontmatter}



\title{Evaluation of Google Translate for  Mandarin Chinese translation using sentiment and semantic analysis}


\author[inst1]{Xuechun Wang} 

\author[inst2]{Rodney Beard}
\author[inst1, inst2]{Rohitash Chandra}

\affiliation[inst1]{Transitional Artificial Intelligence Research Group, School of Mathematics and Statistics, UNSW Sydney, Australia}

\affiliation[inst2]{Centre for Artificial Intelligence and Innovation, Pingala Institute, Nausori, Fiji}
\begin{abstract}
Machine translation using large language models (LLMs) is having a significant global impact, making  communication easier. Mandarin Chinese is the official language used for communication by the government  and  media in  China.  In this study, we provide an automated assessment of translation quality of Google Translate  with human experts using sentiment and semantic analysis. In order to demonstrate our framework, we select  the classic early twentieth-century novel 'The True Story of Ah Q' with  selected Mandarin Chinese to English translations.  We use  Google Translate  to translate  the given text into English and then conduct a chapter-wise sentiment analysis and semantic analysis  to  compare the extracted sentiments across the different  translations.     Our results indicate that the precision of Google Translate differs both in terms of semantic and sentiment analysis when compared to human expert translations.   We find that Google Translate is unable to translate some of the specific words or phrases in Chinese, such as Chinese traditional allusions. The mistranslations may be due to lack of contextual significance and historical knowledge of China.   
\end{abstract}




\end{frontmatter}



\section{Introduction}
\label{sec:sample1}

Natural language processing (NLP) is a sub-field of artificial intelligence that focuses on the interaction between human languages and computers \cite{fanni2023natural} \cite{jones1994natural} \cite{nadkarni2011natural}. NLP is used in various fields of research, such as speech recognition, social media analytics, recommender systems, search engines, machine translation and computational linguistics \cite{chowdhary2020natural}\cite{torfi2020natural}. Machine translation is a branch of NLP that focuses on the automated  translation of one language into another through speech or text \cite{dabre2020survey}.   Computer-assisted translation  (CAT) \cite{craciunescu2004machine} is a software tool designed to help human translators to translate texts faster with better quality \cite{kenny1999cat}. Machine translation systems integrated with CAT tools results in a more productive and accurate  translation process \cite{zaretskaya2015integration}. CAT tools also aim to assist translators in editing and processing translation-related tasks and provides a translation memory system to store source sentences, and translated sentences for future use. CAT tools such as Trados, Déjà Vu, MemoQ, Yicat are predominantly used.  

The range of machine translation models used has changed over time, depending on computational power and technology change. The latest iteration of such models attempts to apply a deep learning model to the language translation problem. Deep learning models such as recurrent neural networks (RNNs)\cite{yu2019review} have been prominent for text and sequence modelling \cite{sutskever2013training}  and their variations form the backbone of text processing and modelling \cite{sutskever2011generating} and machine translation systems\cite{mikolov2011extensions}. RNNs also have been prominent in text classification and generation  \cite{lipton2015critical}. However, canonical models face training challenges and variations such as the long short-term memory (LSTM) \cite{hochreiter1997long}  network has been used to address the problem of learning long-term dependencies. Further alterations led to the development of  the Transformer model \cite{vaswani2017attention}  which employs a multi-head attention mechanism  borrowed from psychology and cognitive science.  Soon afterwards, attempts to train Transformer models with large data corpora led to the development of large language models (LLMs) \cite{zhao2023survey}, including Bidirectional Encoder Representations from Transformers (BERT)  \cite{devlin2018bert,liu2019roberta}, and Generalised Pretrained Transformers (GPT)  \cite{brown2020language} with tools such as ChatGPT, GPT-4 and Gemini. Those models are widely utilised in a wide range of areas addressing problems such as sentiment analysis, text summarising, question-answering and search \cite{dale2021gpt}. However, evaluation of the quality of large language models poses challenges since quantitative evaluation approaches have biases and there are also problems when it comes to explainability and interpretability \cite{chang2024survey}.

Apart from Google Translate, DeepL is also a  widely used machine translation system  \cite{deepl2017}. Yulianto et al. \cite{yulianto2021google} quantitatively compared the performances between Google Translate and DeepL. The research assessed the quality of Google Translate and DeepL for French to English and the results showed that DeepL performed better than Google Translate in terms of accuracy and readability. Moreover, Cambedda et al. \cite{cambedda2021study} studied on a comparative error analysis of a medical translation between DeepL and Yandex for Russian-Italian. This study used the Bilingual Evaluation Understudy (BLEU)\cite{papineni2002bleu} framework to assess the quality of output based on the textual similarities between a machine translation system and a human reference translation. BLEU  was developed by IBM  for evaluating the quality of text which has been machine-translated from one natural language to another. Although BLEU has been prominent for a number of studies \cite{reiter2018structured},  more recently, the use of semantic analysis for evaluating language translation quality has been gaining attention \cite{wieting2019beyond}.



 Popular machine translation tools such as Google Translate \cite{groves2015friend} brought Mandarin Chinese into automatic translation engines including online search, e-learning and social media \cite{ying2021advantages} \cite{guo2016google}, which motivated  studies on the quality of translation. Ling et al. \cite{ling2016differences} investigated the differences in the English and Chinese sentence structure to evaluate the accuracy of English-Chinese and found that the position of clause modifier was a key difference between English and Chinese sentence structures. Lee et al. \cite{lee2011comparative}  compared machine translation and human translations from English to Chinese with post-editing \footnote{Post-editing is a time-consuming process involving manual analysis which can be improved through automation using innovations in NLP. } and reported that machine translation could be useful for saving time and occasionally offered translators thesaurus-like functionality. Although Lee et al. \cite{lee2011comparative} and Ling et al. \cite{ling2016differences} focused on sentence structure and the accuracy of translation, this is a time coning process  manual analysis which can be improved through  discussion of sentiment sectors. 
 
 Recently, Shukla et al. \cite{shukla2023evaluation} presented a study on the  automated analysis of translation assessment using Google Translate and sentiment and semantic analysis for Sanskrit using the Bhagavad Gita. Chandra and Kulkarni \cite{chandra2022semantic} compared translations of Sanskrit to English by three human experts using sentiment and semantic analysis. These studies motivate automation of the evaluation of machine translation engines such as Google Translate for languages such as Mandarin Chinese.


Machine translation has been challenging  for translating between English and Chinese \cite{turner2015machine} \cite{wang2007chinese} due to differences in the linguistic and phonological structure of these languages \cite{perfetti1992reading}.  Taking a linguistic perspective \cite{newmark1981approaches,ziegler2000phonology}, we need to ensure that the evaluation of translations not only takes into account the grammatical accuracy in translations from Chinese to English but also preserves the semantics of the original text.

Automated strategies that include statistical models have taken linguistic perspectives into account \cite{och2004alignment} with statistical machine translation \cite{koehn2009statistical}. In order to evaluate the quality of translation by tools such as Google Translate, we need to select a text that is prominent and has been well-studied with existing expert English translations. We can then compare the Google Translate version of Chinese-English with the expert translation using sentiment analysis, as done for the case of Sanskrit by Shukla et al. \cite{shukla2023evaluation}. In prominent and classic Chinese texts such as novels and other forms of fiction, the expression of emotions is indispensable and readers desire to explore the deep emotions through proper use of language in translated texts. This motivates us to study if  Google Translate can convey the accurate emotions that the original author in Mandarin Chinese intended to express.

In this paper, We present a framework for translation quality analysis using semantic   and sentiment analysis   to  translation of text from Chinese to English. We compare translations of two professional experts   with the translation obtained  from Google Translate. The framework focuses on semantic and sentiment analysis to evaluate the capability of Google Translate to present sentiments (emotions) while maintaining the semantics (meaning). We select prominent English translations  of \textit{The True Story Of Ah Q} (阿Q正传) from professional translators spanning different periods (decades). We use two widely published editions to ensure they are reliable and trustworthy and reduce any translation biases.  

The paper  is organised as follows.  Section 2 presents related work, and Section 3 presents the  Methodology. Section 4 presents the    Results, and Section 5 presents the Discussion and Conclusion.

\section{Related Work}

\subsection{Machine translation}

The early work in machine translation began in the 1930s, which were mechanical, and it wasn't until the development of electronic computers that the possibility of machine translation became a real possibility \cite{hutchins1995machine} \cite{slocum1985survey}. 
NLP was used in World War II for encrypting the secret messages of the German forces and transferring the message to the field commanders and military units of German troops placed in Europe \cite{johri2021natural}.  One of the early NLP domains  developed was machine translation and in the 1950s and 1960s, a multidisciplinary approach with cryptography, statistics, information theory and logic was employed for machine translation research \cite{hutchins1995machine}. 

It is noteworthy that Weaver proposed modern machine translation \cite{weaver1952translation} in the 1950s. During that period of time, even though Weaver's memorandum was of little interest to most linguists or mathematicians, the memorandum still attracted such people as Bar-Hillel from the Massachusetts Institute of Technology and was widely circulated in 1948 \cite{weaver1952translation}. Since then,  machine translation has been popularised that attracted more researchers to the field. Machine translation includes machine learning, statistical and deep learning models \cite{koehn2020neural}. Rule-based machine translation (RBMT) \cite{shiwen2014rule} is based on the syntactic, morphological, contextual and semantic, to perform the machine translation from source language to target languages. This approach was dominated until the late 1980s. The difficulties and problems of rule-based machine translation have arisen because of its time-consuming and complexity. Since then, statistical machine translation has been employed that generally relies on the data-driven approaches.   Moses is a widely used and open-source statistical machine translation toolkit \cite{koehn2007moses} that incorporates a number of language models, including a neural translation model. Moreover, it provides a number of functions such as data preprocessing, data-training, and fine-tuning in the toolkit. 

\subsection{Mandarin Chinese translation analysis }


Early efforts in Mandarin Chinese machine translation began with Wang et al. \cite{wang1973chinese} who developed the segmentation of Chinese character strings into words, conversion of traditional Chinese and simplified Chinese. Along with the development of Chinese computational linguistics research in the 21st century, numerous studies in the field of Mandarin Chinese translation have rapidly increased. For example, Wang et al. \cite{wang2007chinese} introduced a set of syntactic rules that utilize systematic word order differences between Chinese and English, and further transform Chinese sentences to English with much closer word order. Wang et al. \cite{wang2007chinese} reported that their reordering approach has improved translation quality with relatively reordering errors. Deng et al. \cite{deng2017translation} presented an empirical investigation that explored the divergences between English and Chinese by using a parallel tree bank to semi-automatically recognise divergences(such as lexical encoding, transitivity, absence of function words, category mismatch, reordering, dropped elements, and structural paraphrase) between English and Chinese.
Furthermore, to evaluate the quality of machine translation from Chinese to English, Jia et al. \cite{jia2019post} investigated the differences between from-scratch translation and post-editing of Google  Translate using both qualitative and quantitative analysis, examining aspects such as accuracy and fluency from multidimensional perspectives. Post-editing is a time-consuming process involving manual analysis of text, which can be improved through innovations in NLP. Additionally, Ma et al. \cite{ma2022generative} proposed a survival generative adversarial network-based machine translation method to evaluate the short sequence machine translation from Chinese to English. These substantial studies expand the scope of machine translation from Chinese Mandarin to English.






%

\subsection{Google Translate}

On the other hand, since the advent of Google Translation in 2007, the core of Google Translate was to translate one language to another language including but not limited to words, sentences, paragraphs and articles via text and speech. Google Translate used statistical models and proved that results are better than rule-based system.  Maulidiyah et al. \cite{maulidiyah2018use} reported that in 2010, Google Translate was ranked first place among other search engines such as Yahoo, Baidu, Bing, and occupied 84.65\% of the total market. Given  the high penetration rate of Google Translate, researchers have explored the accuracy and precision of Google Translate.  Li et al. \cite{li2014comparison} compared the Google Translate with expert (human) translation through its formality and cohesion.  The correlation with formality is measured by Coh-Metrix \cite{graesser2014coh}, and Linguistic Inquiry and Word Count (LIWC) \cite{pennebaker2007expressive}. Moreover, the correlation with cohesion is measured by Latent Semantic Analysis (LSA) \cite{landauer2007lsa} and Content Word Overlap (CWO) \cite{mcnamara2014automated}. Li et al. \cite{li2014comparison} concluded that both translations had high correlation in terms of both formality and cohesion. 

Moreover, Patil et al. \cite{patil2014use} investigated the accuracy of Google Translate in medical communication with multiple languages, and reported  57.7\% of correct  translations and  42.5 \% incorrect; hence  Google Translate could not be used for medical processes due to poor accuracy. Cromico et al. \cite{cromico2015translation} evaluated the quality of the translation of English-Indonesian by utilising scientific articles, and reported that Google Translate was less readable, less accurate, and less acceptable for the target language. Therefore, although Google Translate has made life more convenient, the issues of inaccuracy, low precision and readability were still shortcomings of Google Translate. However, due to its popularity among the public and the large number of Chinese speakers, the quality of Google Translate for Chinese texts is worth exploring.


\subsection{Mandarin Chinese}

The Mandarin Chinese language is spoken by more than 1.1 billion people in the world with the majority of speakers residing in China \cite{julian2020most}.  Mandarin Chinese is the second most spoken language in the world after English \cite{wiki:languages}. From a linguistic perspective, following conventions in linguistic typology, both Chinese and English are considered to be subject-verb-object languages \cite{greenberg1963some}. For example, 我 (subject of the sentence) 吃 (verb of the sentence) 苹果（object of the sentence). In English, the sentence would be translated as "I (subject) eat (verb) apple (object)".   Another example would be "我"  (subject of the sentence) 是 (copula) 中国人 (object of the sentence). In English, the sentence would be translated as "I (subject) am (copula) Chinese (object)". The two examples follow SVO (subject-verb-object) order, where the subject goes first, followed by the verb (copula) and then the object. Thus, the order of sentences from Chinese and English could be categorised as SVO in the perspective of typology. However, English and Chinese differ in terms of grammar rather than morphology. For instance, the use of tense in most Western languages compared with the use of incomplete action using the particle 了, and the use of particles which are also used in Western languages; e.g. {\em doch} in German, {\em si} in French {\em eh}  in English but are more fundamental in the different Chinese dialects, e.g. the question particle 吗. Furthermore，the Mandarin Chinese is a language without tense morphology \cite{lin2006time}. For example, in English, we can state,  "I had my breakfast." Note that 'had' as a verb in this sentence is phrased in the past tense. However, the sentence in Chinese would be "我吃了早饭"(wo chi le zao fan), '吃'(eat) as a verb in this case does not change tense because of time.  However, the aspectual particle '了' suffixed to the verb is used for past and future interpretation. Collart et al. \cite{collart2021processing} stated that English is a "tense prominent" language where tend to place an event in time tense, whereas Chinese is a "aspect prominent" language that refer to the time aspect through aspectual particles. Thus, as a language without tense morphology, Mandarin differs from English in how it expresses time. Instead of using verb tenses, Mandarin employs aspectual particles such as "了"(-le), "过"(guo) as suffixes to its verbs for temporal interpretation.


\subsection{Sentiment and Semantic Analysis}

Word embedding in  NLP encodes text numerically as a dense vector for machine and deep learning models \cite{abubakar2022sentiment, li2018word}.  
 Initially, the text sequences were represented by simple methods such as one hot encoding and later, different word-embedding methods were developed  \cite{mikolov2013efficient,ghannay2016word}. For instance, 
Mikolov et al. \cite{mikolov2013efficient} introduced the Word2Vec word embedding model in 2013   which achieved highly accurate word vectors via training a simple neural network. The model efficiently enabled the computation of high-dimensional word vectors with lower computational complexity. On the other hand, Pennington et al. \cite{pennington2014glove} introduced the \textit{GloVe} model (global log-bilinear regression model) that enables the training of nonzero elements in a word-word concurrence matrix for effectively leveraging information. 


Semantic analysis in NLP   investigates the meaning of languages, which is useful for understanding the representation of texts from English to other languages \cite{salloum2020survey} enabling cross-lingual semantic analysis  \cite{csenel2018generating}. 

The use of a similarity measure for the comparison of different sets of data is required for the evaluation of information retrieval processes, clustering, and classification models \cite{santisteban2015unilateral}. The Jaccard similarity score is used for measuring the similarity and difference between two data samples that include text phrases represented as strings \cite{zhang2021semantic}. The cosine similarity computes the cosine similarity between two strings that can also feature vector embeddings\cite{rahutomo2012semantic}. 

Sentiment analysis is used for the analysis of the text by evaluating people's opinions,   emotions,   and impressions through data available in social media platforms, news media,  and journals \cite{wankhade2022survey}.   Sentiment analysis is a growing field and is employed prominently in social media as a tool to assess impressions of users for product advertising \cite{taboada2016sentiment}. Sentiment analysis has been used for a wide range of applications, including election modelling and forecasting \cite{chandra2021biden}, COVID-19 social media analysis \cite{chandra2023analysis}, and news media analysis \cite{chandra2024large}. Our study employs sentiment analysis to compare sentiments captured by Google Translate and human translation.

\section{Methodology}
\label{sec:sample1}
\subsection{Description of dataset}

\textit{The True Story of Ah Q} （啊Q正传）\cite{karki2010true} is a classic short novel that depicts the real life of Chinese society in 1911. Lu Xun as a social realist satirised feudal culture and the social, political and economic circumstances of the time. The story contains numerous expressions of emotions including empathy, humour, hopelessness and irony. In the context of  The True Story of Ah Q,   the difficulties and challenges Ah Q faced symbolised the pressures of feudal thinking and an emerging capitalist society in the early twentieth century in China. Lu Xun's works have entered the global market and been translated into many languages, translations give the work a new life and translation styles can vary based on their understanding and diverse perspectives on the literature \cite{huan2023corpus}.

 The True Story of Ah Q  consists of nine chapters and was first published as a serial in \textit{Chenbaofukan} (晨报副刊) on the fourth of December, 1921. It has since been reproduced in his short story collection called \textit{Call To Arms}（呐喊） in 1923. Lu Xun is considered to be probably the greatest Chinese writer of the 20th century, and his literary works have been broadly translated by numerous scholars \cite{von2012translating}. Because of his iconic status in Chinese literature, we selected one of his best-known pieces for analysis. In addition, we selected two English translation works of \textit{The True Story Of Ah Q} to evaluate translation quality and for model training. 

One of the selected English translations was done by Yang Xianyi and his wife, Gladys Taylor Yang. It was published by the Chinese Communist government-run Foreign Languages Press in Beijing in 1953. Their works brought Lu Xun's literary work onto the world stage and introduced and popularised Chinese works in Western countries to improve understanding of Chinese culture during that period. Gladys Yang has expressed three thoughts on translation: being creative in translation, conforming to the readers' consciousness, and adapting to the current policy \cite{gangguistudy}. Moreover, her translation style takes into account both \textit{foreignisation} and \textit{domestication} \cite{gangguistudy}. Foreignisation is a type of translation strategy that retains some of the foreignness of original texts \cite{yang2010brief}. Domestication is designated for fluent style in order to help the target language reader better understand the content. The translation style should be thought of as considering both native readers and foreign readers, and to increase the reader's understanding of the Chinese cultural background of a text.


Another English translation of \textit{The Real Story of Ah Q} was translated by Julia Lovell and published by Penguin Classic in November 2009 \cite{wang2014interview}. Besides \textit{The Real Story of Ah-Q and Other Tales of China}, Julia Lovell has translated a number of other books from Chinese, such as \textit{A Dictionary of Maqiao} from Han Shaogong, \textit{I Love Dollars and Other Stories of China} from Zhu Wen, \textit{Serve the People!} from Yan Lianke. Julia Lovell has extensive experience translating Chinese texts, which is why we chose her translations for comparison. In \textit{The Real Story of Ah-Q and Other Tales of China}, Julia Lovell translated Lu Xun's stories into a single volume as a collection. However, in our study, we focused only on the novella of \textit{The Real Story of Ah-Q}. Julia Lovell has stated that she translated these texts literally without embellishment, which may have been closer to the spirit of the original \cite{wang2014interview}. In other words, this is a direct translation.

Our study also considered other well-known translations, such as those by William Lyell \cite{Lu_Lyell_1976}. However, we have not used it in this study and recommend for future studies. 

\subsection{Data extraction and processing}

We extracted all nine chapters from the original text and then translated them to English using the Google Translate application programmer interface (API) \footnote{API: \url{https://translation.googleapis.com/language/translate/v2}}. We extracted all the verses by scraping web text data. 
We pre-processed data with the following steps:
\begin{itemize}
 
  \item  Opening a printable document format (PDF) file, and reading the content of each chapter.

  \item Removing chapter number and chapter name and annotation numbering from each chapter.

  \item  Separating each chapter given as paragraphs to a single line so that we can implement sentiment analysis based on sentences rather than a paragraph.

 
\end{itemize}

After pre-processing data, we published three versions of translations in GitHub \footnote{Github: \url{https://github.com/sydney-machine-learning/translationanalysis-Mandarin/tree/main}}. 

\subsection{BERT for sentiment and semantic analyses}

The  BERT  \cite{devlin2018bert} model is a pre-trained large language model that employs the Transformer deep learning model \cite{vaswani2017attention} that incorporates an attention mechanism in an LSTM-based model. The Transformer model processes text using a left-to-right architecture, whereas the BERT model generalises this to a bidirectional process.  The BERT model takes advantage of knowledge from a large corpus of data that includes Wikipedia and Bookus databases and features billions of trainable parameters. A pre-trained model essentially provides a trained model that can be downloaded and further refined using task-specific datasets. In our study, we refine the BERT-based model using task-specific data for sentiment analysis. 


\subsection{Framework}

\begin{figure*}[ht]
    \centering
    \includegraphics[width=\textwidth]{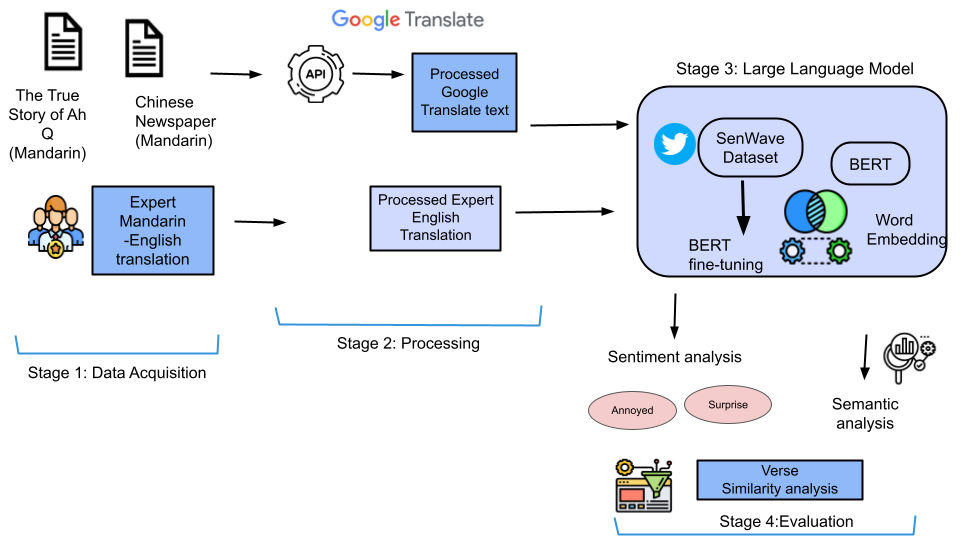}
    \caption{A framework representing stages of processes of sentiment and semantic analysis for Mandarin to English, comparing Google Translate and human experts.}
    \label{fig:framework}
\end{figure*}


We present a framework that features sentiment and semantic analysis by comparing the Google Translate version with the translations from Mandarin to English of two expert translations. We note that our framework has been adopted from Chandra et al. \cite{chandra2022semantic} and Shukla et al. \cite{shukla2023evaluation} (Figure \ref{fig:framework}) which evaluated translations of Sanskrit to English. Our framework detects sentiments based on sentences in each translation for chapter-to-chapter comparison between human experts and the Google Translate version.  


In Stage 1, we first scraped texts from the original texts of Lu Xun found on the internet (url\url{https://www.marxists.org/chinese/reference-books/luxun/03/012.htm}), and the translations of Yang Xianyi and Gladys Taylor Yang \cite{Lu_Yang_Yang_1990}, as well as another translation from Julia Lovell \cite{Lu_Lovell_2009}. Afterwards, in Stage 2, we  convert PDF files of Mandarin texts and human expert translations into text files to further apply the data processing. Then, we read the original text files into the Google Translate application processor interface (API) and publish the corresponding English translations  and their processed versions to the GitHub repository \footnote{\url{https://github.com/sydney-machine-learning/translationanalysis-Mandarin/tree/main/Data_set}}. We split the texts by chapters, and then we saved chapter-wise files   as 'Google dataset', 'Julia dataset' and 'Hsien dataset'(Hsien is the surname of Xianyi), respectively. 

In Stage 3, we use a BERT-based model to implement sentiment analysis of each sentence and provide a summary of the chapter-wise sentiment. We provide multi-label sentiment classification in our framework, which means each verse could have more than one sentiment. We also train our BERT model by utilising the Senwave dataset, which includes 10 different emotions(\textit{Optimistic}, \textit{Thankful}, \textit{Empathetic}, \textit{Pessismistic}, \textit{Anxious}, \textit{Sad}, \textit{Annoyed}, \textit{Denial}, \textit{Official report}, \textit{Joking})  \cite{yang2020senwave}. The SenWave dataset \cite{yang2020senwave} features Tweets from March 1 2020 to May 15 2020 and contains 10,000 human labelled Tweets which enabled it to be utilised in various studies, particularly for refining pre-trained language models. For example, Chandra et al. \cite{chandra2022semantic} evaluated the Bhagavad Gita translation (Sanskrit to English) using sentiment analysis with a fine-tuned BERT model based on the SenWave dataset.


In addition, in Stage 3, we also employ the MPNet-based sentence embedding model \cite{song2020mpnet} to quantitatively observe the similarities between the three translations. Our framework aims to obtain results on how similar human experts' translations are compared to Google Translate. 

In stage 4, we visualise the sentiment analysis in separate plots for each translation. We construct cumulative sentiment analysis plots for each translation (accumulating the sentiments of each chapter) corresponding to each translation. These multi-label classification plots intend to provide insights into similarities and differences of sentiments captured for  Google Translate and human translations. Furthermore, we present cosine similarity for comparing selected pairs of three translations. It is further used to compare the most and least semantically similar verses among three translations to evaluate the semantic quality of Google Translate translation in more detail.

Our framework combines quantitative evaluation of Google Translate texts via sentiment and semantic analysis. We then provide qualitative analysis by an expert assessment of the  translations.

\subsection{Experimental and technical setup}


 We utilised the pre-trained BERT-base uncased model and fine-tuned the model on the SenWave dataset \cite{yang2020senwave} with a dataset of 10,000 tweets by using a batch size of 1 and 4 training epochs. Since the sentiment 'official report' is related to COVID-19, we deleted this sentiment in our work as it is not related to our texts. Moreover, the latest SenWave dataset removed\footnote{The latest version of Senwave Dataset: \url{https://github.com/gitdevqiang/SenWave/blob/main/labeledtweets/labeledEn.csv}} the 'Surprise' in their dataset; therefore we also removed 'Surprise' from our framework to avoid bias.  Moreover, we implement regularisation by using a dropout layer \cite{srivastava2014dropout} with a probability of 0.3 dropout to address the over-fitting problem in model training. Furthermore, we use a linear activation layer at the output later for predicting 9 sentiments present in the SenWave dataset.   We use the pre-trained BERT model for comparing selected translations of \textit{The True Story Of Ah Q} via sentiment analysis. 


\section{Results}

\subsection{Data Analysis}

\begin{figure}[htbp!]
    \centering
    \begin{subfigure}[b]{0.45\textwidth}
        \centering
        \includegraphics[width=\textwidth]{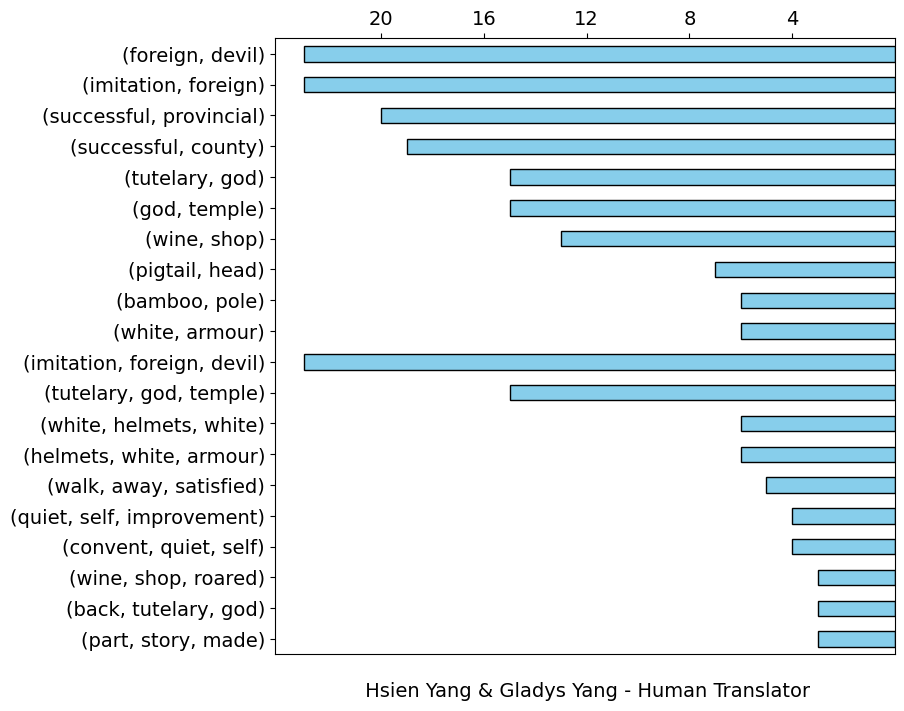}
        \caption{The Real Story of Ah Q by Xianyi Yang and Gladys T Yang}
        \label{fig:plot1}
    \end{subfigure}
    \hfill
    \begin{subfigure}[b]{0.45\textwidth}
        \centering
        \includegraphics[width=\textwidth]{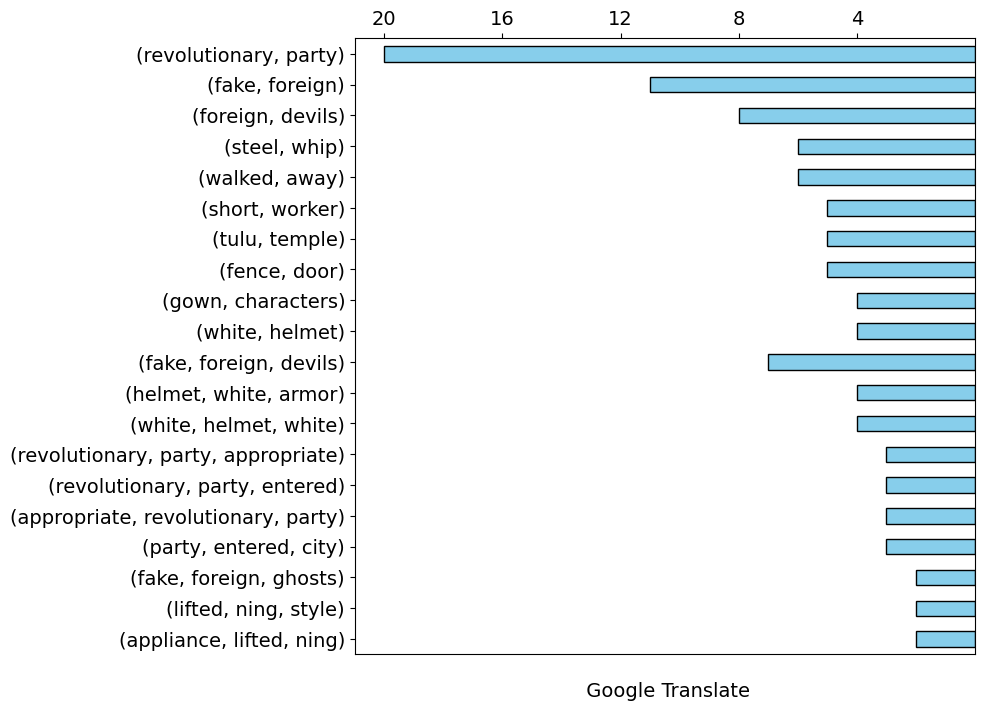}
        \caption{The Real Story of Ah Q by Google Translate}
        \label{fig:plot2}
    \end{subfigure}
    \hfill
    \begin{subfigure}[b]{0.45\textwidth}
        \centering
        \includegraphics[width=\textwidth]{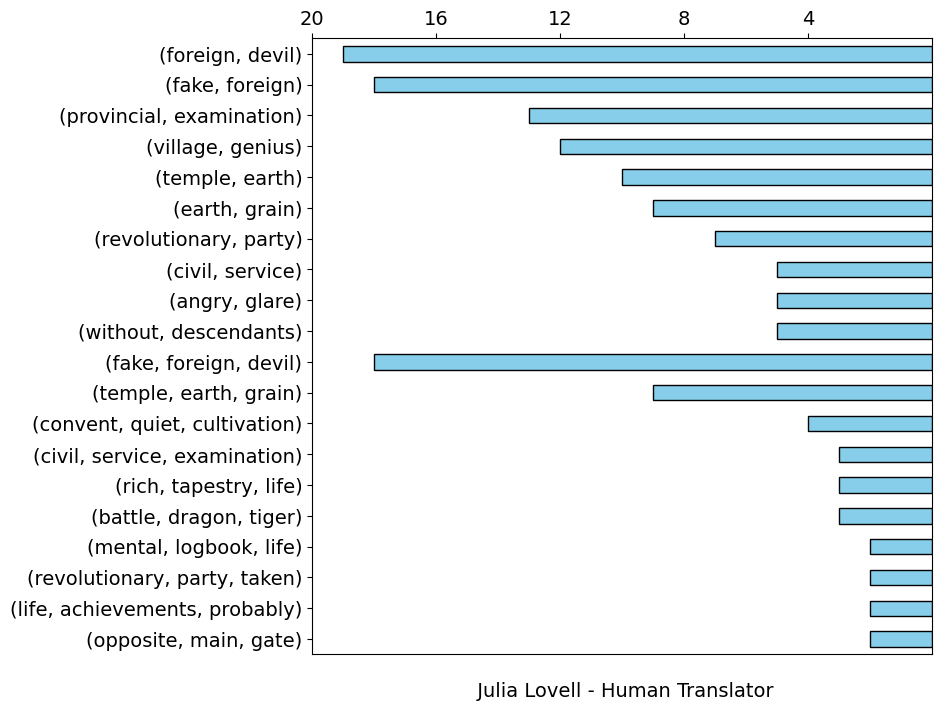}
        \caption{The Real Story of Ah Q by Julia Lovell}
        \label{fig:plot3}
    \end{subfigure}
    \caption{Framework of top 10 bi-grams and top 10 trigrams for different translations.}
    \label{fig:n_grams}
\end{figure}

N-grams are NLP methods used for \cite{robertson1998applications} extracting adjacent characters, symbols or words from a given text.  We removed stop words and utilised n-grams (bigrams and trigrams) for each translation, to get an understanding about the difference vocabulary.  We present the top 10 ranked bigrams and trigrams for different translations of \textit{The Real Story of Ah-Q} in Figure \boxedref{fig:n_grams}

We review the bigrams and trigrams of Google Translate as shown in  Figure \boxedref{fig:plot2}; where [revolutionary, party] and [fake, foreign, devils] are the most frequently mentioned consecutive words in texts for bigrams and trigrams, respectively. This reflects that some elements such as, 'rebellious', 'devil' are recurring themes in the Google Translate version. Moreover, it also highlights that some words such as 'party', 'killing', 'helmet',  imply that the translation's tone is somewhat negative. Moreover, the bi-grams and trigrams of the translation by Xianyi Yang and Gladys Taylor Yang (Figure \boxedref{fig:plot1}), emphasise words like  [foreign, devil] and [imitation, foreign, devil]. These are the most frequently mentioned consecutive words for bigrams and trigrams correspondingly. In  Figure \boxedref{fig:plot1}, words such as "devil", "god", "temple" are frequently captured, and hence, the translation by Xianyi Yang and Gladys Yang seem to contain  religious and historical themes. In addition, the bigrams and trigrams of the translation by Julia Lovell (Figure \boxedref{fig:plot3}), feature the bi-grams [foreign, devil], and trigrams [fake, foreign, devil].  We note  that the most frequently mentioned consecutive words in bigrams and trigrams overlap. Moreover, "Fake," "temple," "earth," "grain," "revolutionary," and "party" are also frequently used words in Figure \boxedref{fig:plot3}. Note that the highest frequency consecutive words used by Figure \boxedref{fig:plot1}(Xianyi Yang and Gladys Yang) and Figure \boxedref{fig:plot3} (Julia Lovell) are very similar to each other. This provides additional support that the two translations by three human expert translators are relatively similar in their use of keywords.

\begin{figure*}[htbp!]
    \centering
    \includegraphics[width=\textwidth]{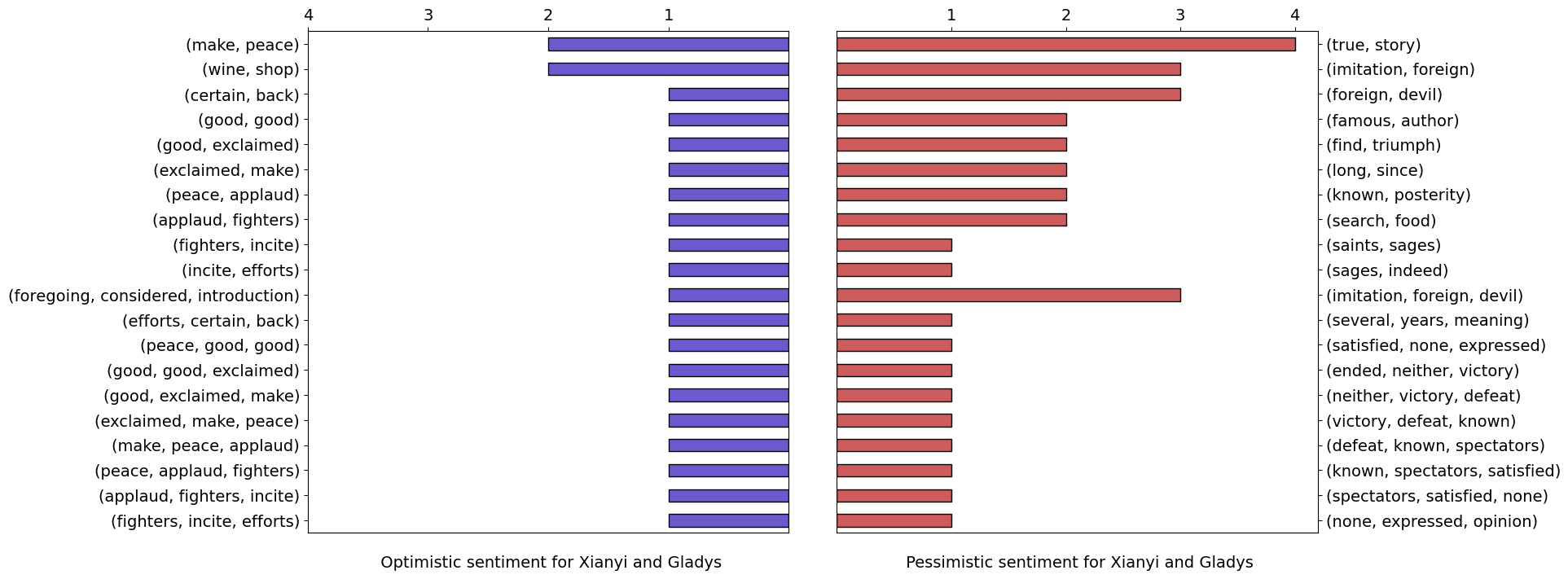}
    \caption{Top 10 optimistic and pessimistic bigrams and trigrams from Xianyi Yang and Gladys Yang's translation }
    \label{fig:hsien_sentiment_ngrams}
\end{figure*}

\begin{figure*}[htbp!]
    \centering
    \includegraphics[width=\textwidth]{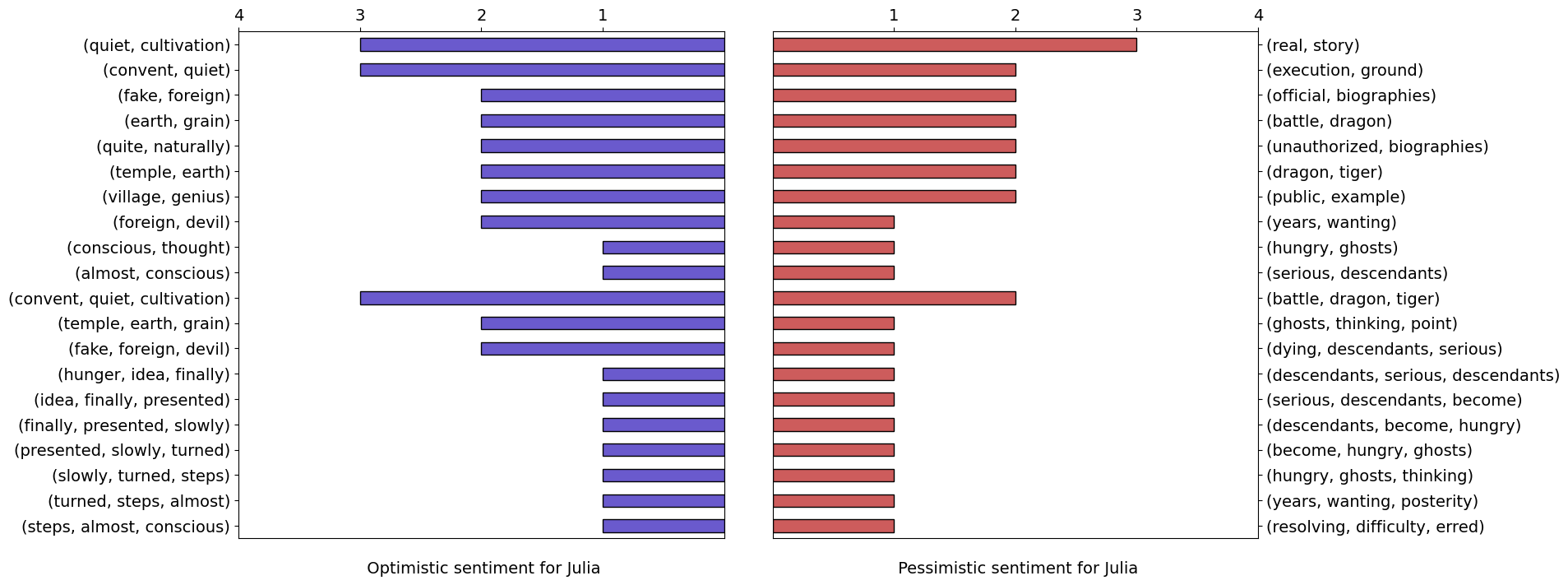}
    \caption{Top 10 optimistic and pessimistic bigrams and trigrams from Julia Lovell's translation }
    \label{fig:julia_sentiment_ngrams}
\end{figure*}

\begin{figure*}[htbp!]
    \centering
    \includegraphics[width=\textwidth]{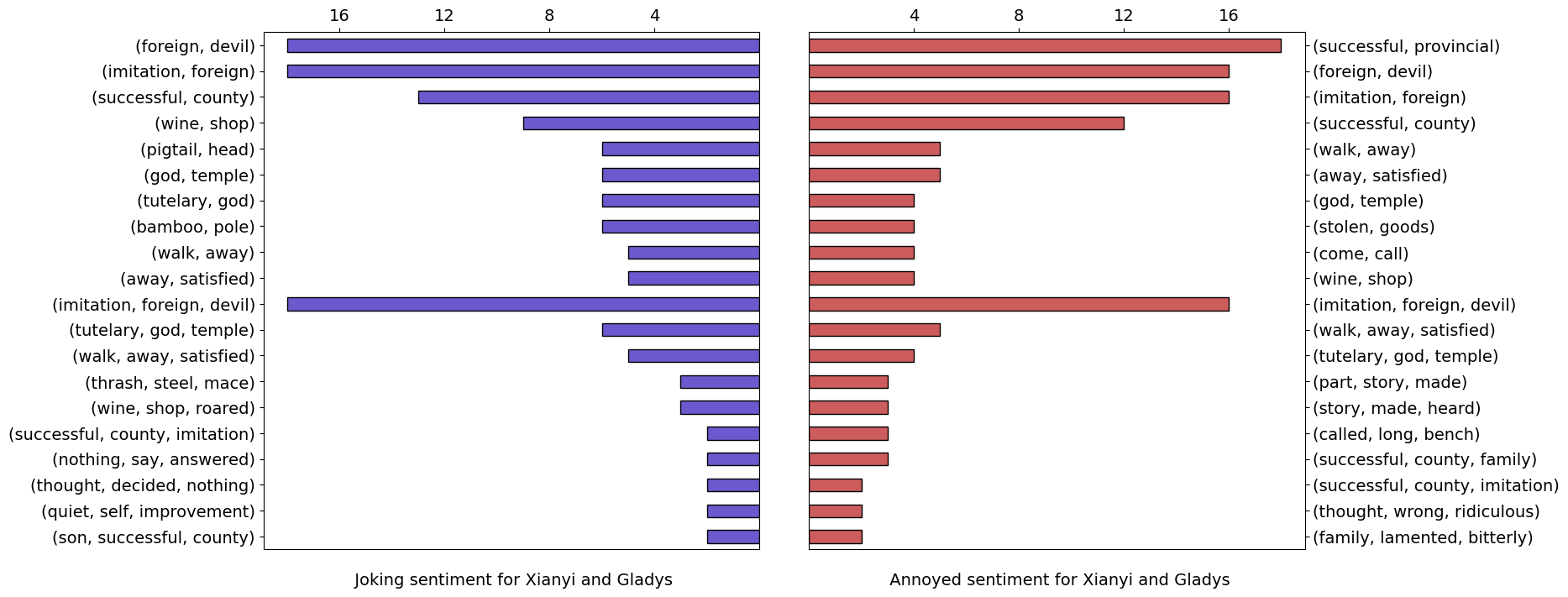}
    \caption{Top 10 joking and annoyed bigrams and trigrams from Xianyi Yang and Gladys Yang's translation }
    \label{fig:hsien_sentiment_joyking_ngrams}
\end{figure*}

\begin{figure*}[htbp!]
    \centering
    \includegraphics[width=\textwidth]{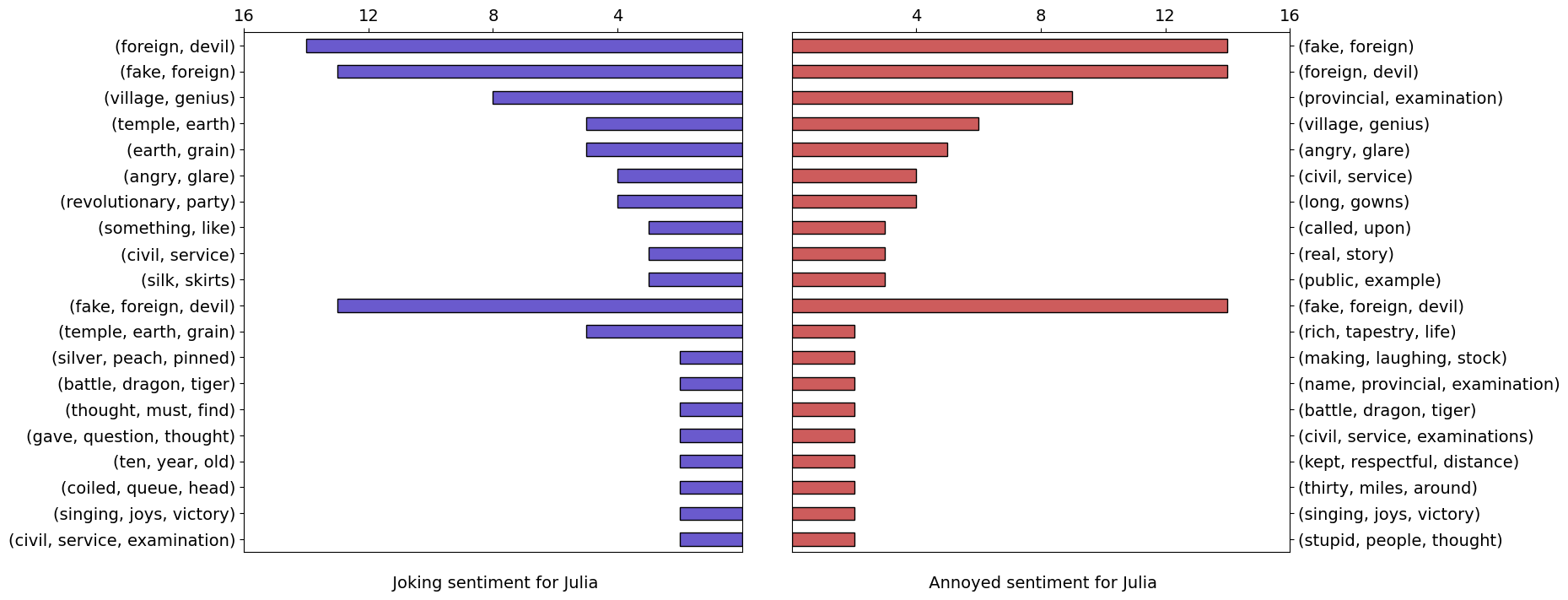}
    \caption{Top 10 joking and annoyed bigrams and trigrams from Julia Lovell's translation }
    \label{fig:julia_sentiment_joking_ngrams}
\end{figure*}


In comparing the Google Translate with the human expert versions  \boxedref{fig:n_grams}, we find   commonly mentioned phrases  such as "devil", "temple". In the meantime, from Figure \boxedref{fig:n_grams}, [foreign, devil] is the most frequently occurring bigram for both human translators. Although [foreign, devil] appears in bigrams of the Google Translate version, the most frequent bigram for Google Translate is [revolutionary, party]. 

The common keywords for three translations are "foreign", "devil", "temple", "fake", those words lleadto a gative themetic patterns. Moreover, it is noteworthy to indicate that the second most frequent permutation in bigrams and the most frequently occurring trigrams permutation for both human translators have similar themes, but are different in their use of keywords("imitation" and "fake"). Hence, although Google Translate can extract some common keywords in its translation, the overall high-frequency keywords differ from those translations translated by human experts.

\subsection{Sentiment analysis}

\begin{figure}[htbp!]
    \centering
    \includegraphics[width=\linewidth]{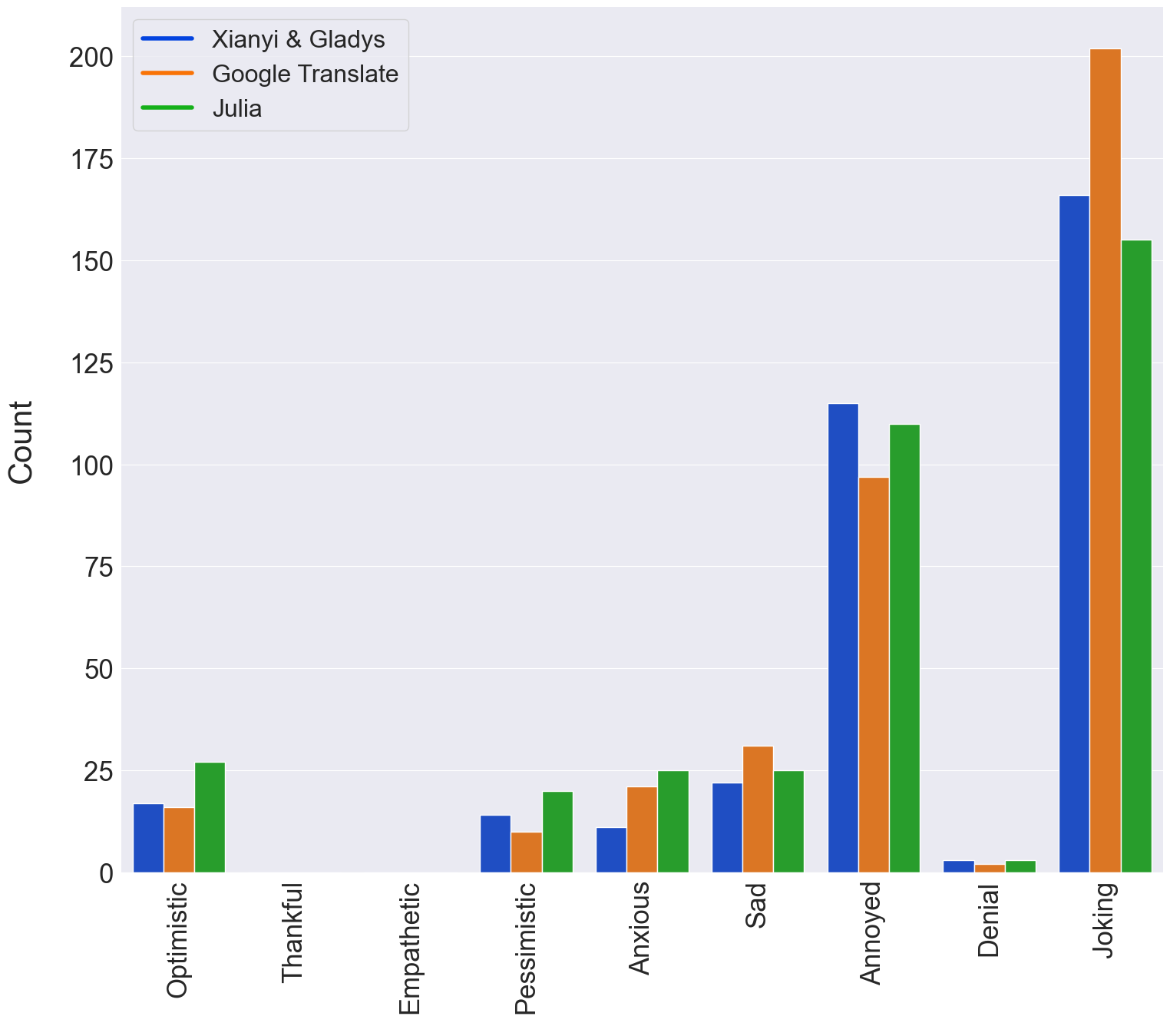}
    \caption{Cumulative sentiment analysis across all chapters}
    \label{fig:combo_combo}
\end{figure}

Next, we present a data visualisation for verse-by-verse sentiment analysis for three different translations using the refined BERT model. We first show the top 10 ranked optimistic and pessimistic bigrams and trigrams of two human expert translations in Figures \boxedref{fig:hsien_sentiment_ngrams} \boxedref{fig:julia_sentiment_ngrams}.

 Figure \boxedref{fig:hsien_sentiment_ngrams} represents top 10 optimistic  and pessimistic   bigrams and trigrams across nine chapters for Xianyi and Gladys translation. Furthermore, Figure \boxedref{fig:julia_sentiment_ngrams} represents the top 10 optimistic and pessimistic  bigrams and trigrams across nine chapters for Julia's translation. In Figure \boxedref{fig:hsien_sentiment_ngrams} and \boxedref{fig:julia_sentiment_ngrams},  [true, story] and [real, story] are the most mentioned pessimistic words in the two different translations. We note that "true story" and "real story"  in the title of the two translations have the same meaning. Since they are extracted as top-ranked pessimistic permutations, it further highlights the gloomy tone of the story for both translations. Moreover, based on the bigrams, the optimistic most frequently used words are [make, peace] and [wine, shop] for Xinyi and Gladys, whereas the most common  words are [quiet, cultivation] and [convent, quiet] for Julia. Additionally, in terms of predicted optimistic leading words for trigrams, [foregoing, considered, introduction] and [efforts, certain, back] are the most frequently occurring permutations for Xianyi and Gladys, then [convent, quiet, cultivation] and [temple, earth, grain] are the most common permutations for Julia. On the other hand, the predicted pessimistic leading words in bi-grams are [true, story] and [imitation, foreign] for Xinyi and Gladys, whereas the most common words are [real, story] and [execution, ground] for Julia. Furthermore, in terms of predicted common words for tri-grams. [imitation, foreign, devil] is the most common for Xianyi and Gladys, whereas [battle, dragon, tiger] are the most common words for translation by Julia.  We note that [foreign, devil] is listed in both two translations. However, it is listed in optimistic bigrams for Xinyi and Glady's translation version and  also present in the pessimistic bigrams for Julia's translation. A potential reason could be the context in which verses are phrased in the different texts.

 \textcolor{black}{In Figure \ref{fig:hsien_sentiment_joyking_ngrams} and Figure \ref{fig:julia_sentiment_joking_ngrams}, we observe that [foreign, devil] and [successful, provincial] is leading permutation in joking and annoyed bi-grams for Xianyi and Gladys, while [foreign, devil] and [fake, foreign] is leading words in joking and annoyed bi-grams for Julia. Moreover, by evaluating joking and annoyed tri-grams for all translations, it is interesting to note that [imitation, foreign, devil] is leading words in both joking and annoyed sentiments for Xianyi and Gladys, and [fake, foreign, devil] is leading words in both joking and annoyed sentiments for Julia. It further illustrates that the expression in terms of joking and annoyed by two human experts translation are generally similar. Thus, it gives a message, that the cumulative counts of ’Annoyed’ and ’Joking’ sentiments in the two translations shows no huge differences in Figure \ref{fig:combo_combo}.}

\begin{figure*}[htbp!] 
    \centering
    \includegraphics[width=0.75\textwidth]{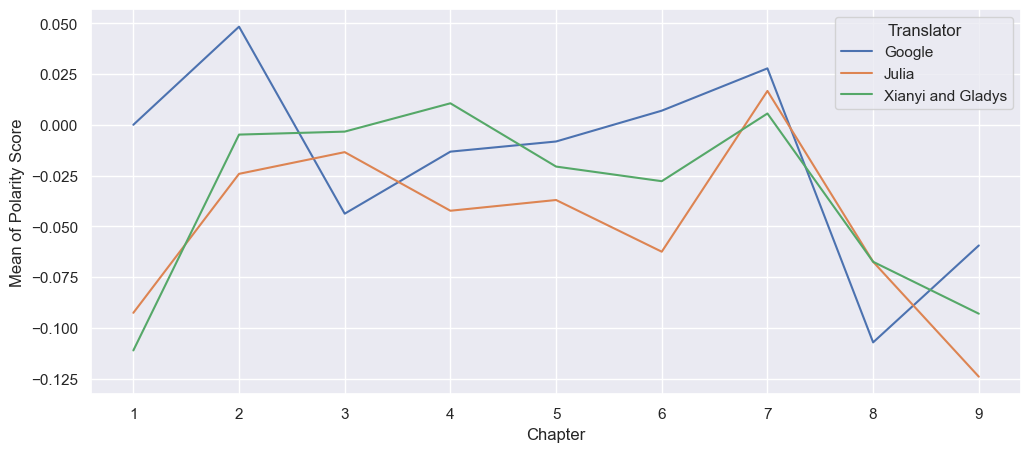}
    \caption{Mean polarity score across nine chapters with three translations Google Translate, Julia, Xianyi and Gladys).}
    \label{fig:polarity_mean} 
\end{figure*}

\begin{figure*}[htbp!] 
    \centering
    \includegraphics[width=0.75\textwidth]{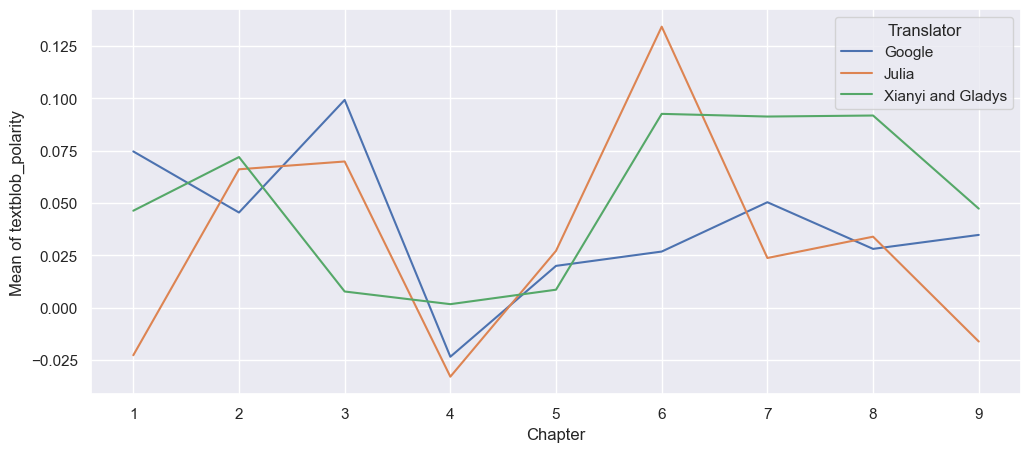}
    \caption{Mean polarity score obtained by Textblob across nine chapters with the threetranslations (Google Translate, Julia,  Xianyi and Gladys). }
    \label{fig:polarity_textblob} 
\end{figure*}

 \textcolor{black}{Additionally, we explore the core sentiment of each chapter leading towards either positive or negative sentiments, thus we visualise the sentiment polarity score of each chapter in Figure \ref{fig:polarity_hsien}. We categorise the value of each sentiment as either 1 or -1, depending on whether it is a positive sentiment or a negative sentiment. We assign sentiments "optimistic" with a positive score of + 2, "thankful" with a score of + 3 and "joking" with a score of + 1 as they are positive sentiments. In contrast, we assign sentiments "pessimistic" with a score of -4, "anxious" with a score of -2, "sad" with a score of -3, "annoyed" with a score of -1 and "denial" with a score of -5. Then, we assign "Empathetic" with a score of 0 since it could be treated as either positive or negative sentiments. By observing Figure \ref{fig:polarity_mean}, it is interesting to note that the mean of polarity score across all chapters are mostly negative (except Google Translate translation). The trend of two human experts translation represent similar features, which both start with an increasing trend, remain relatively steady trend in middle chapters, then show an upward trend from chapter 6 to chapter 7 and end with a downward trend. This further demonstrates that the two human experts translations have a similar overall understanding for the original text in sentiment perspective. However, the mean of polarity for Google Translate translation has different trend compared with two human experts translation. For example, the translation by Google Translate shows a obvious downward trend from chapter 2 to chapter 3, whereas the feature is not observed from two human translations. The polarity score of Google Translate translation deviates from the two human experts translations. Figure \ref{fig:polarity_textblob} is used a textual data processing library in Python known as TextBlob. The range of polarity score in TextBlob is given from -1 to 1. By comparing mean of polarity score by first method and mean of polarity score by TextBlob, the range of the mean polarity score by TextBlob is higher than the mean polarity score by the first method. Moreover, in Figure \ref{fig:polarity_textblob}, the mean of polarity scores across three translations are different. We notice that both Google Translate translation and Julia translation has a lowest point in chapter 4 whereas the translation by Xianyi and Glady does not exhibit this feature. It is interesting to note that the translations by two human experts show a downward trend from chapter 8 to chapter 9 in both Figure \ref{fig:polarity_mean} and Figure \ref{fig:polarity_textblob}. This reflects that Ah Q involved in the revolution and ultimately becomes a victim for the revolution and being executed in chapter 9, leading to a tragic end.}

\begin{figure*}[htbp!]
    \centering
    \begin{subfigure}[b]{0.45\linewidth}
        \centering
        \includegraphics[width=\linewidth]{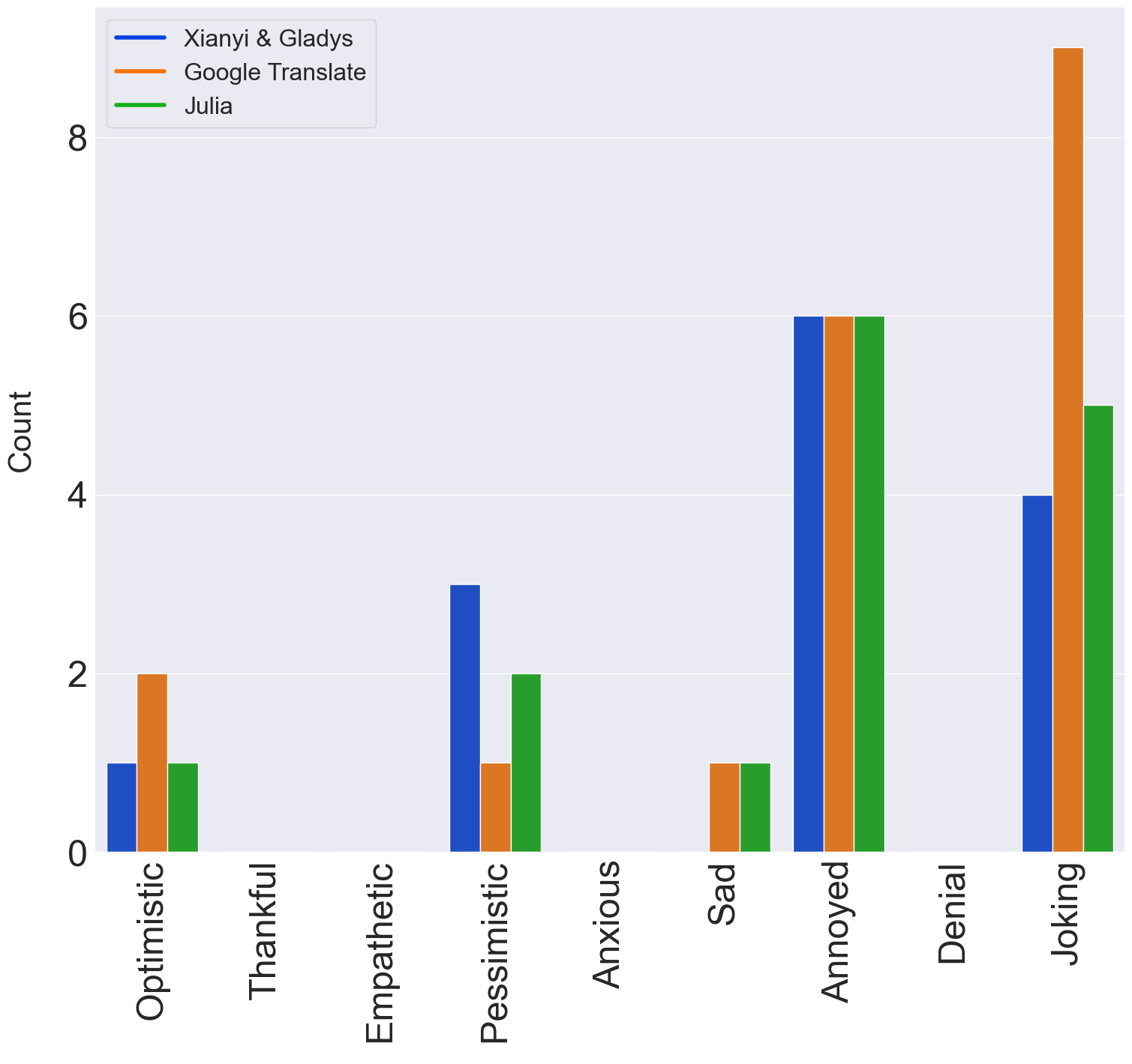}
        \caption{Chapter-1}
        \label{fig:combo_1}
    \end{subfigure}
    \hfill
    \begin{subfigure}[b]{0.45\linewidth}
        \centering
        \includegraphics[width=\linewidth]{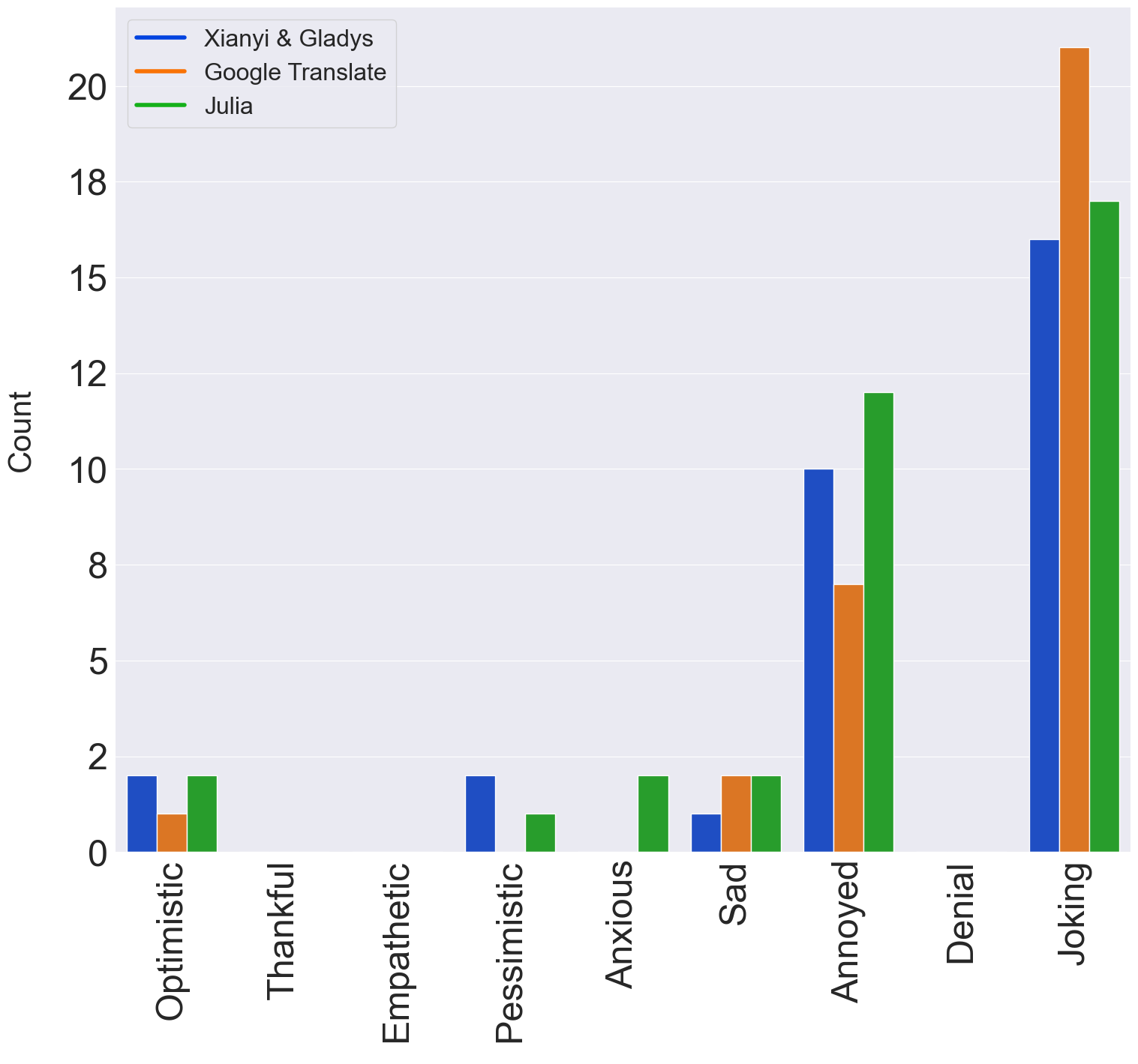}
        \caption{Chapter-2}
        \label{fig:combo_2}
    \end{subfigure}
    \hfill
    \begin{subfigure}[b]{0.32\linewidth}
        \centering
        \includegraphics[width=\linewidth]{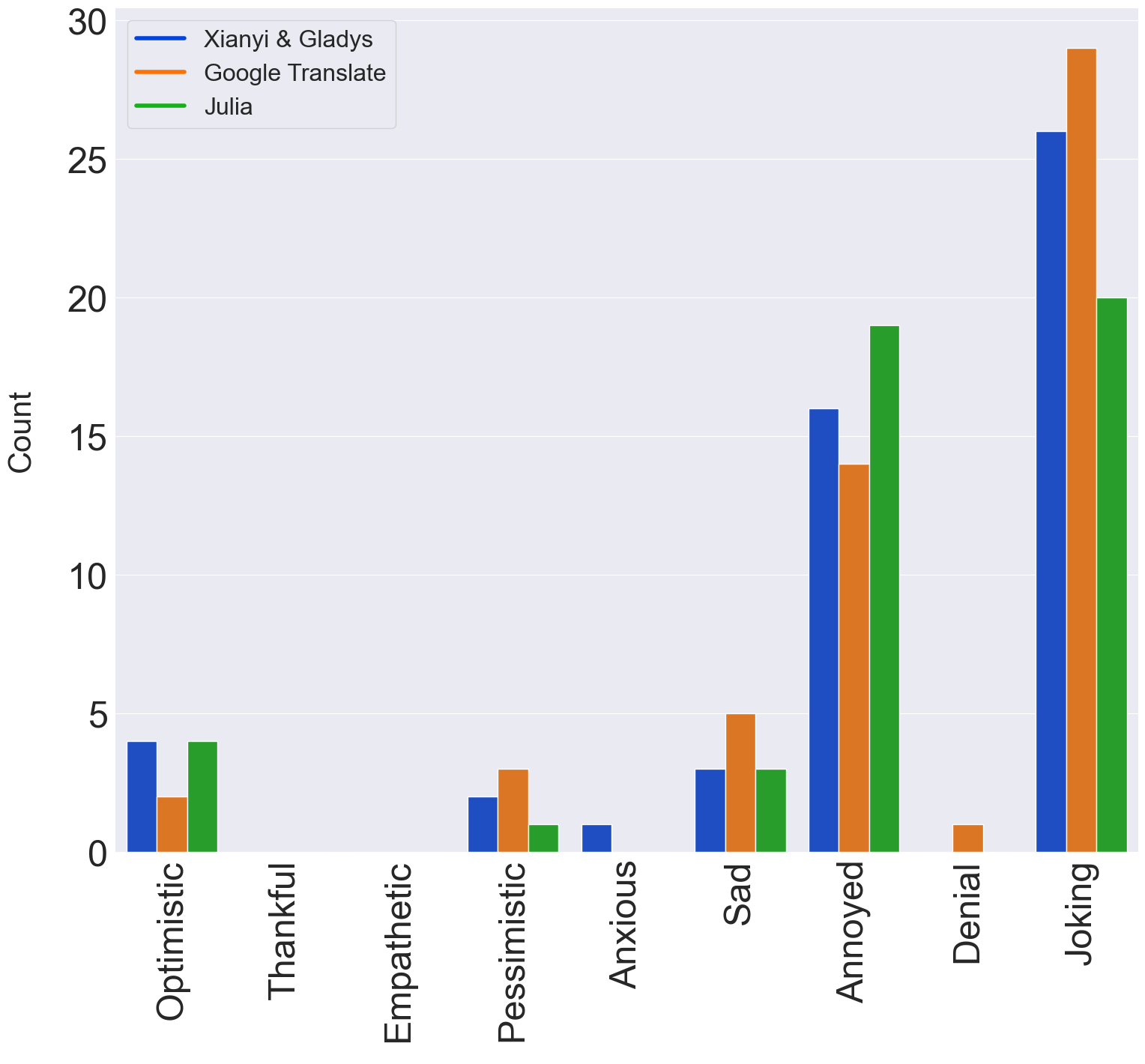}
        \caption{Chapter-3}
        \label{fig:combo_3}
    \end{subfigure}
    \hfill
    \begin{subfigure}[b]{0.32\linewidth}
        \centering
        \includegraphics[width=\linewidth]{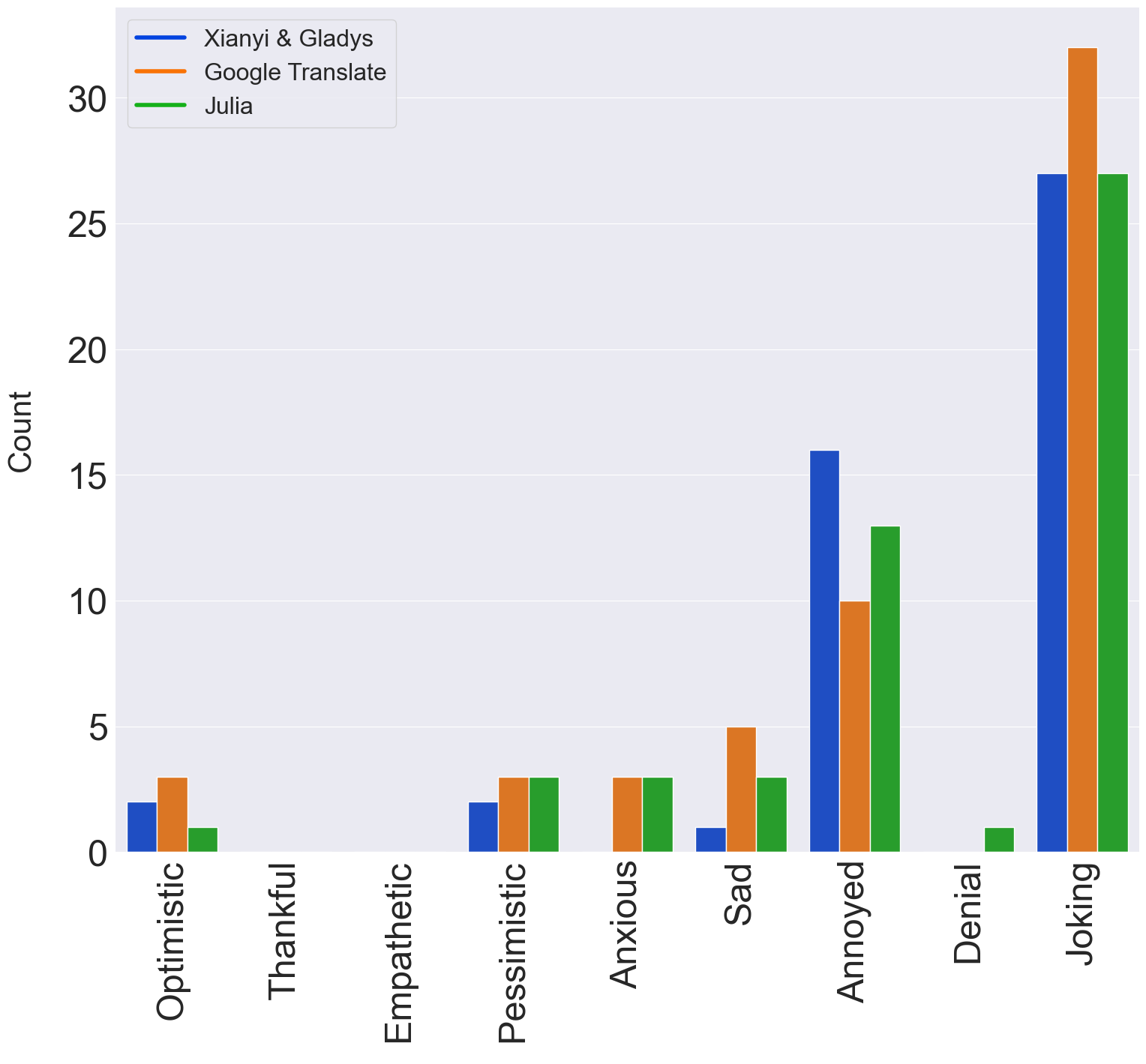}
        \caption{Chapter-4}
        \label{fig:combo_4}
    \end{subfigure}
    \hfill
    \begin{subfigure}[b]{0.32\linewidth}
        \centering
        \includegraphics[width=\linewidth]{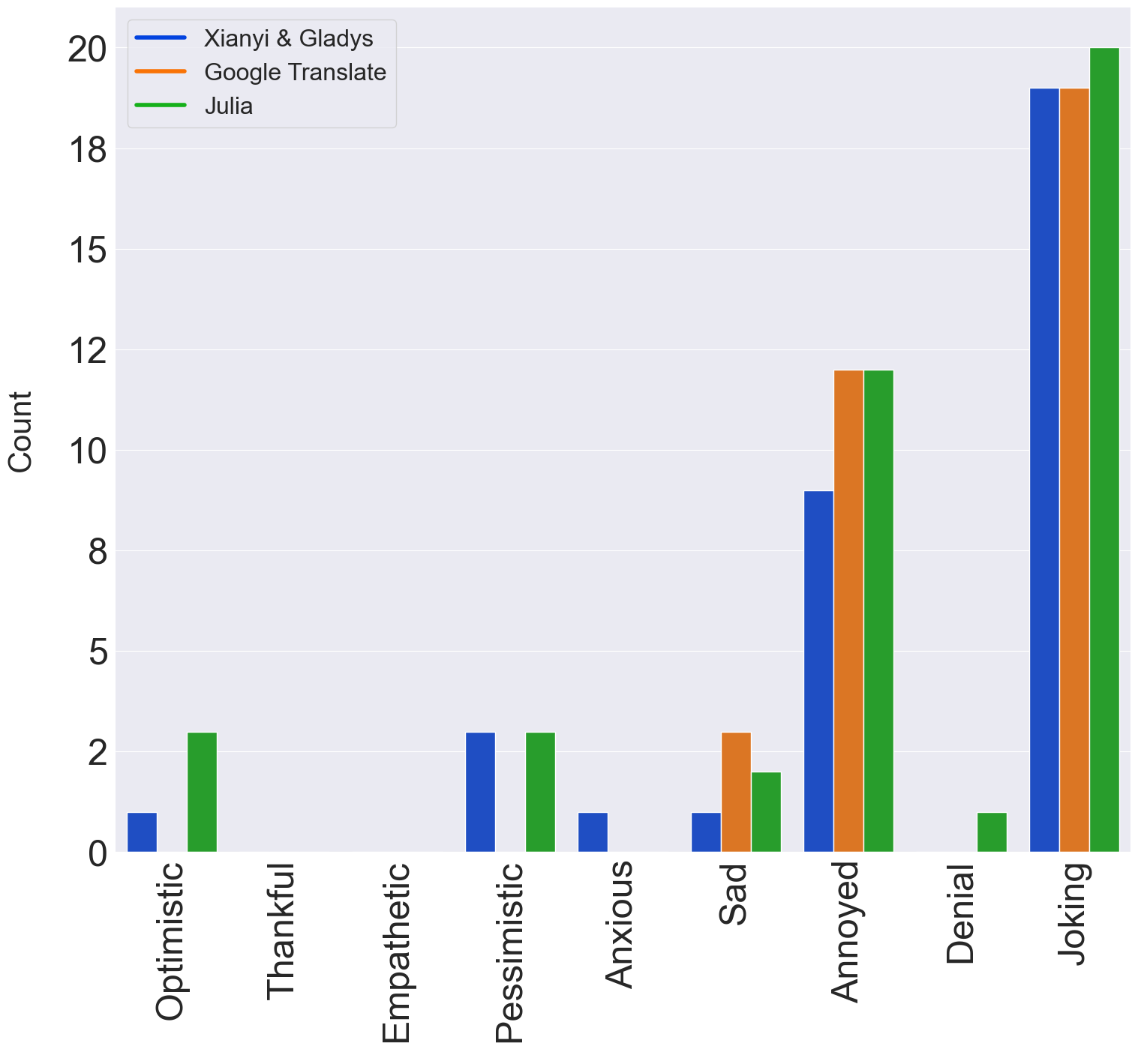}
        \caption{Chapter-5}
        \label{fig:combo_5}
    \end{subfigure}
    \hfill
    \begin{subfigure}[b]{0.32\linewidth}
        \centering
        \includegraphics[width=\linewidth]{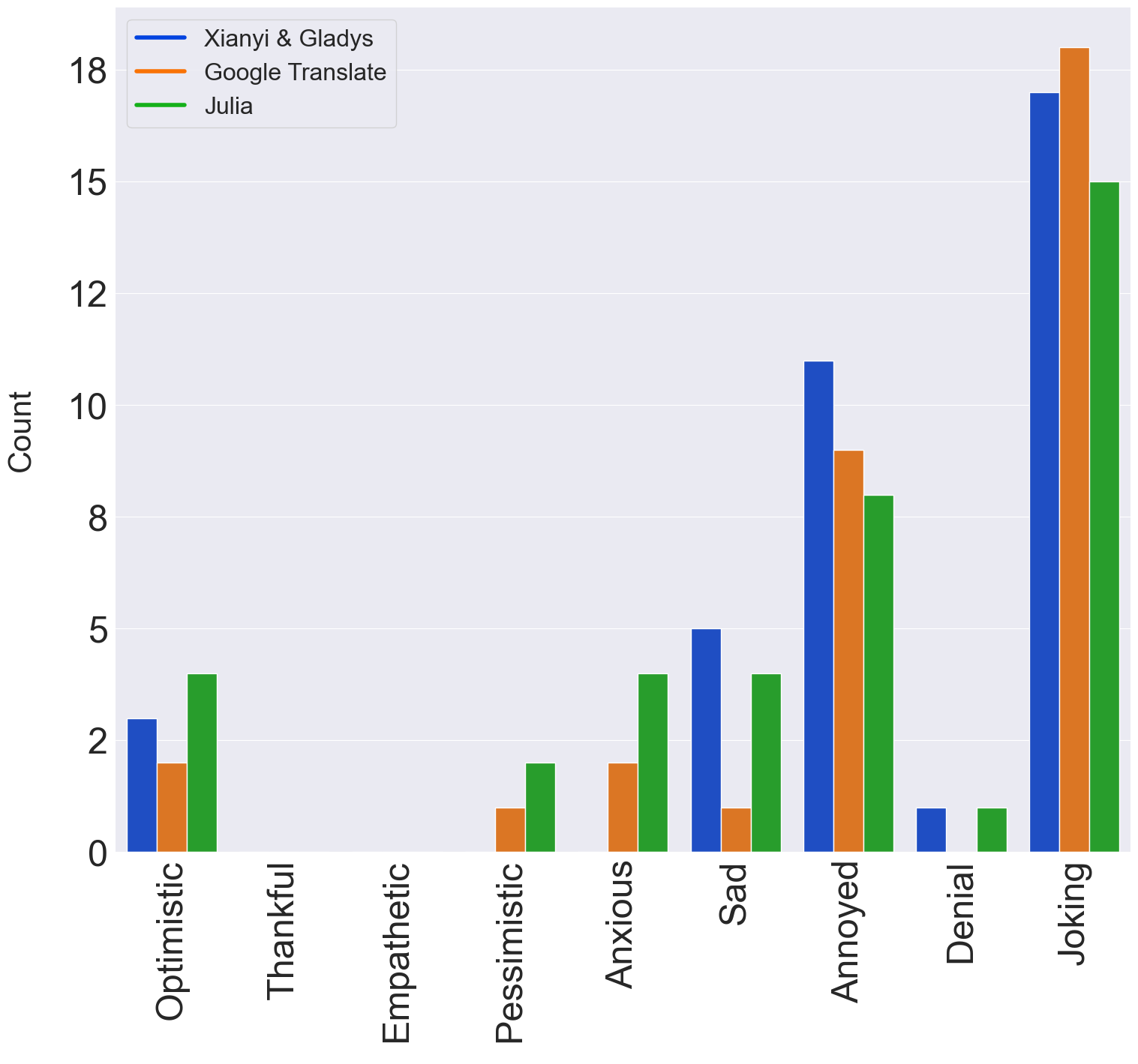}
        \caption{Chapter-6}
        \label{fig:combo_6}
    \end{subfigure}
    \hfill
    \begin{subfigure}[b]{0.32\linewidth}
        \centering
        \includegraphics[width=\linewidth]{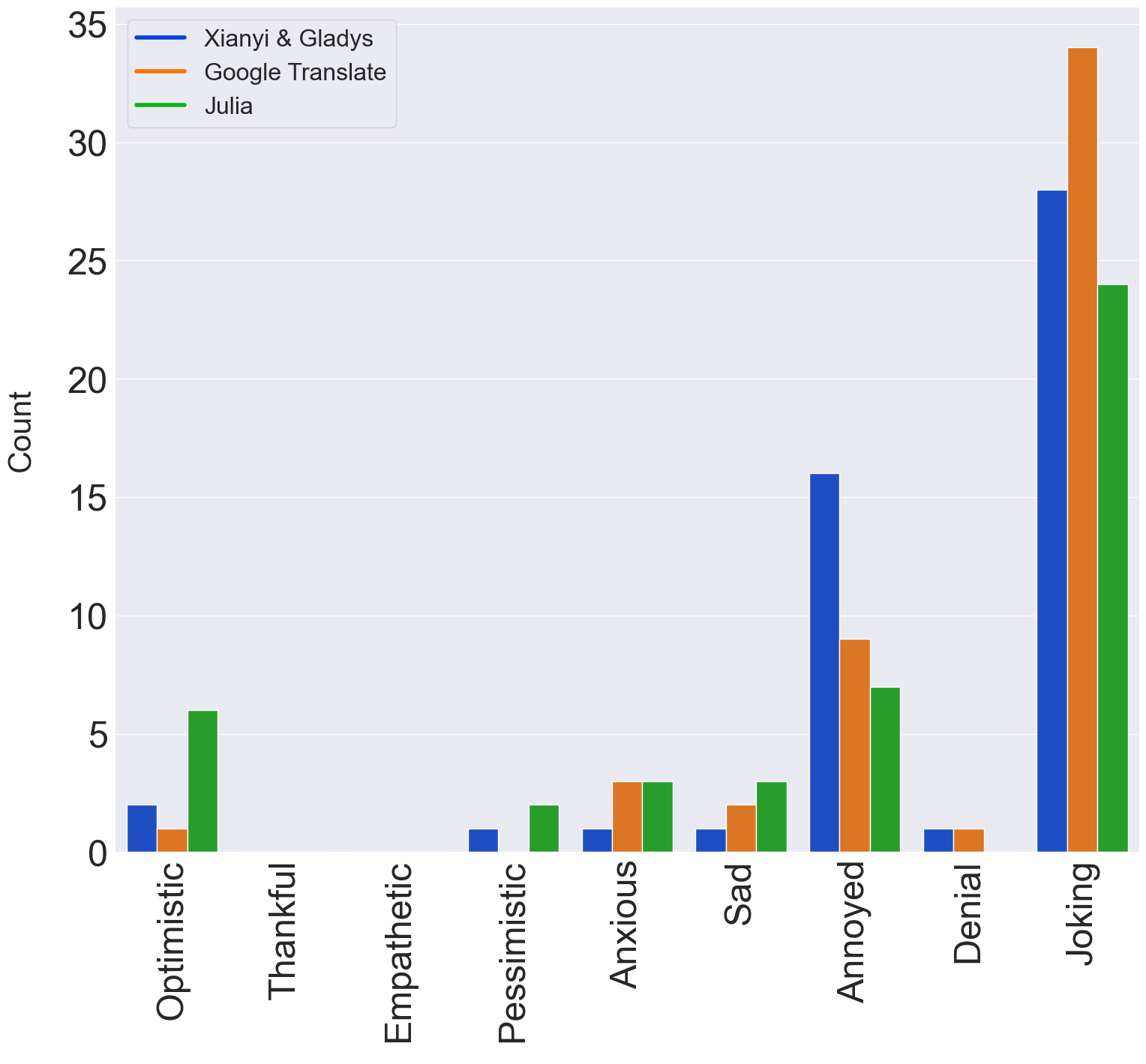}
        \caption{Chapter-7}
        \label{fig:combo_7}
    \end{subfigure}
    \hfill
    \begin{subfigure}[b]{0.32\linewidth}
        \centering
        \includegraphics[width=\linewidth]{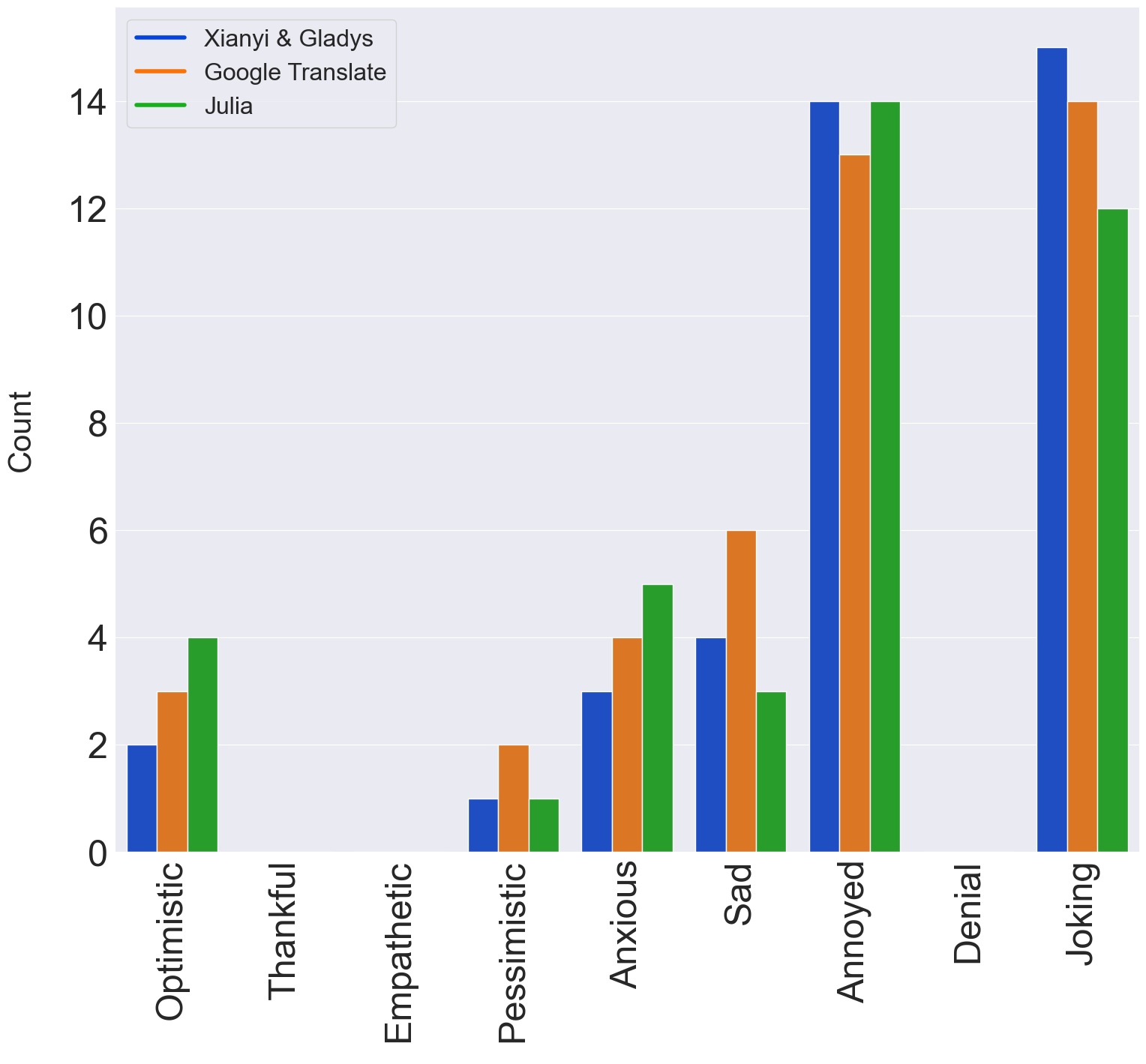}
        \caption{Chapter-8}
        \label{fig:combo_8}
    \end{subfigure}
    \hfill
    \begin{subfigure}[b]{0.32\linewidth}
        \centering
        \includegraphics[width=\linewidth]{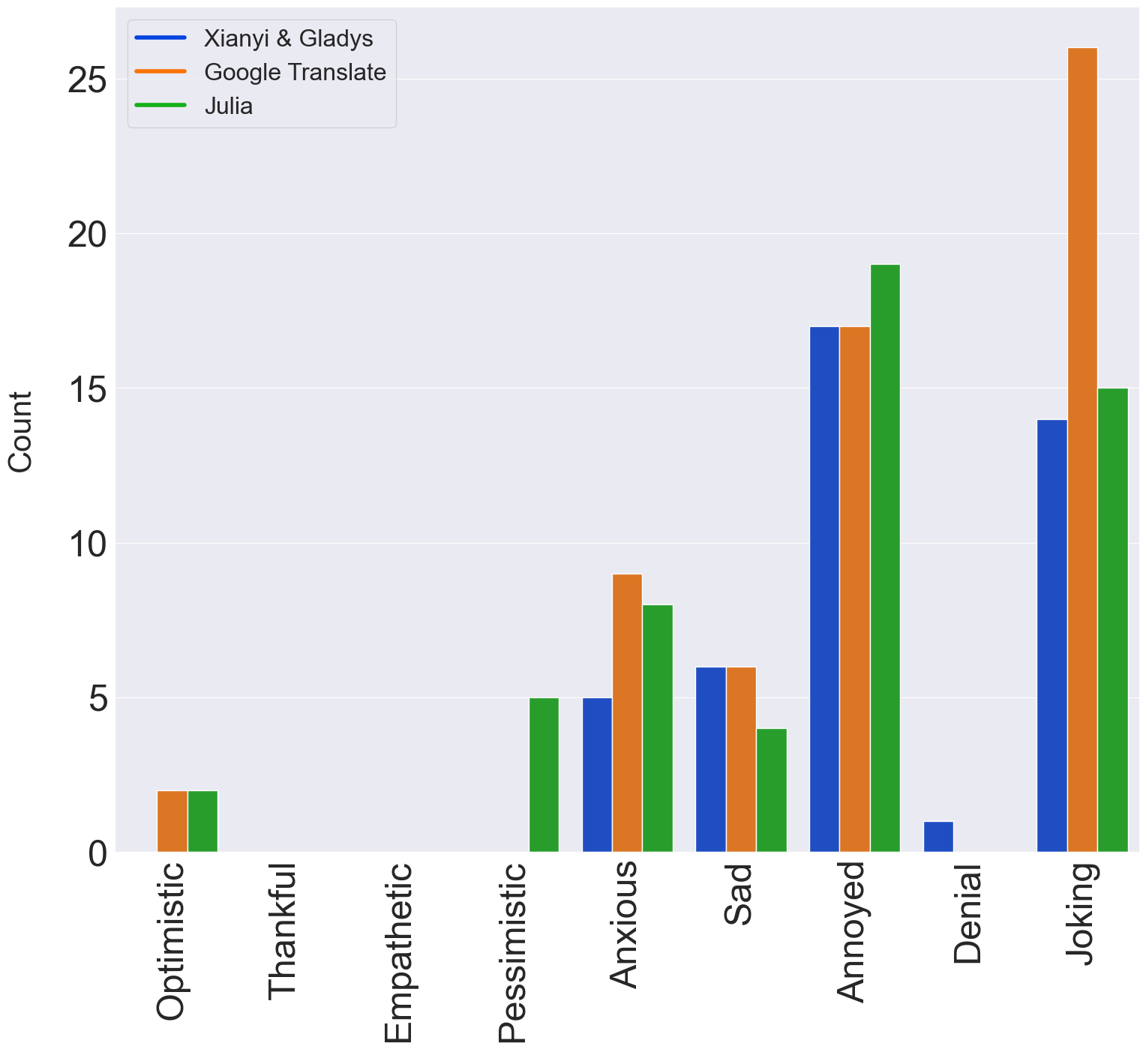}
        \caption{Chapter-9}
        \label{fig:combo_9}
    \end{subfigure}
    \caption{Chapter-wise sentiment analysis from Chapter1 to Chapter9.}
    \label{fig:combo_n}
\end{figure*}

We then present sentiment analysis of entire text along with the chapter-wise sentiment analysis from Chapter 1 to Chapter 9 (Figure \boxedref{fig:combo_combo}\boxedref{fig:combo_n}). In the Figure \boxedref{fig:combo_combo}, we observe that \textit{Thankful}, \textit{Empathetic} and \textit{Denial} are least expressed sentiments across three translations. We note that \textit{Thankful} and \textit{Empathetic} are not expressed across all nine chapters in three translations. Thus, there are only seven sentiments that are captured in \textit{The Real Story of Ah-Q}. Moreover, Figure \boxedref{fig:combo_combo} also demonstrates that Google Translate's version dominant sentiments are \textit{sad}, \textit{joking} which does not necessarily refer to a joke but also captures humour. Xianyi and Gladys   dominant sentiment is \textit{annoyed}, whereas Julia Lovell's  dominant sentiments are  \textit{optimistic} and \textit{pessimistic}. We note that the cumulative sentiment \textit{denial} counts from Xianyi and Gladys, and Julia versions are the same. Furthermore, it highlights that sentiment \textit{annoyed}" is under-represented by Google Translate, whereas  \textit{joking} is over-represented by Google Translate. Moreover, the cumulative sentiment analysis plot also depicts that \textit{annoyed} and \textit{joking} are the dominant sentiments for all three translations. Thus, the three translations place equal emphasis on the main sentiments, however, the importance of the dominant sentiments in the translations by human experts is relatively closer to each other.

\begin{figure}[htbp!]
    \centering
    \begin{subfigure}[b]{0.4\textwidth}
        \centering
        \includegraphics[width=\textwidth]{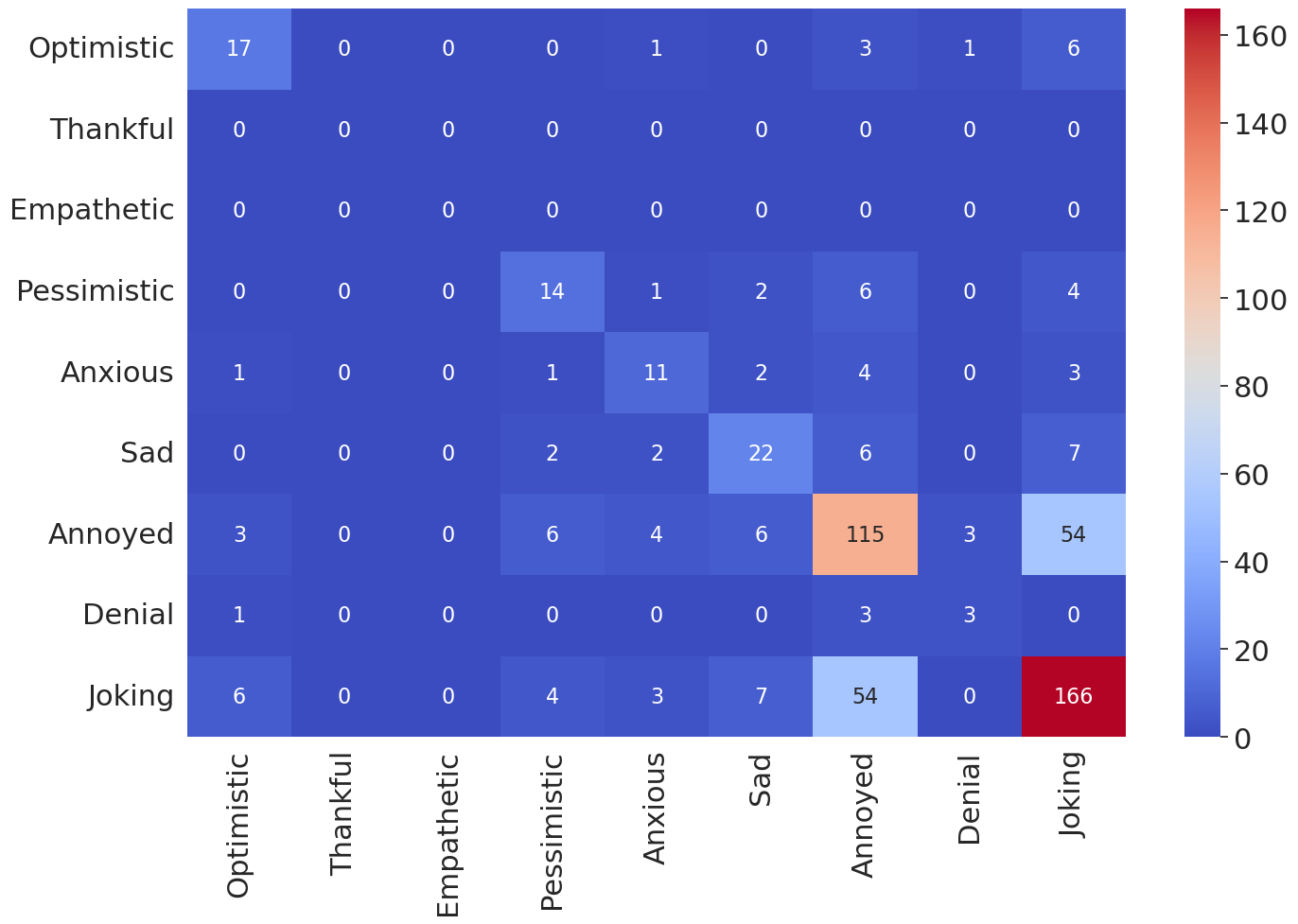}
        \caption{Xianyi and Gladys}
        \label{fig:heatmap1}
    \end{subfigure}
    \hfill
    \begin{subfigure}[b]{0.4\textwidth}
        \centering
        \includegraphics[width=\textwidth]{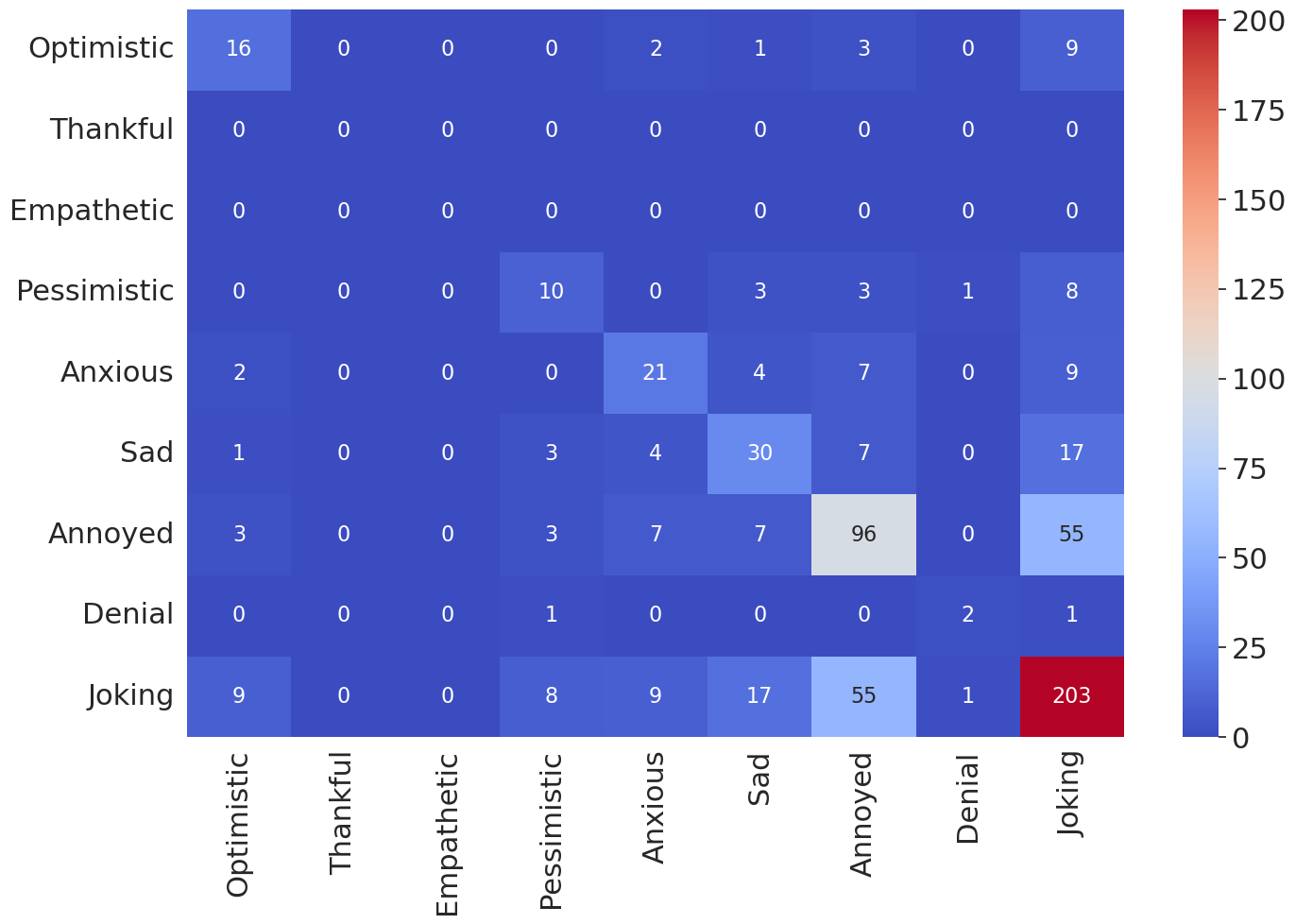}
        \caption{Google Translate}
        \label{fig:heatmap2}
    \end{subfigure}
    \hfill
    \begin{subfigure}[b]{0.4\textwidth}
        \centering
        \includegraphics[width=\textwidth]{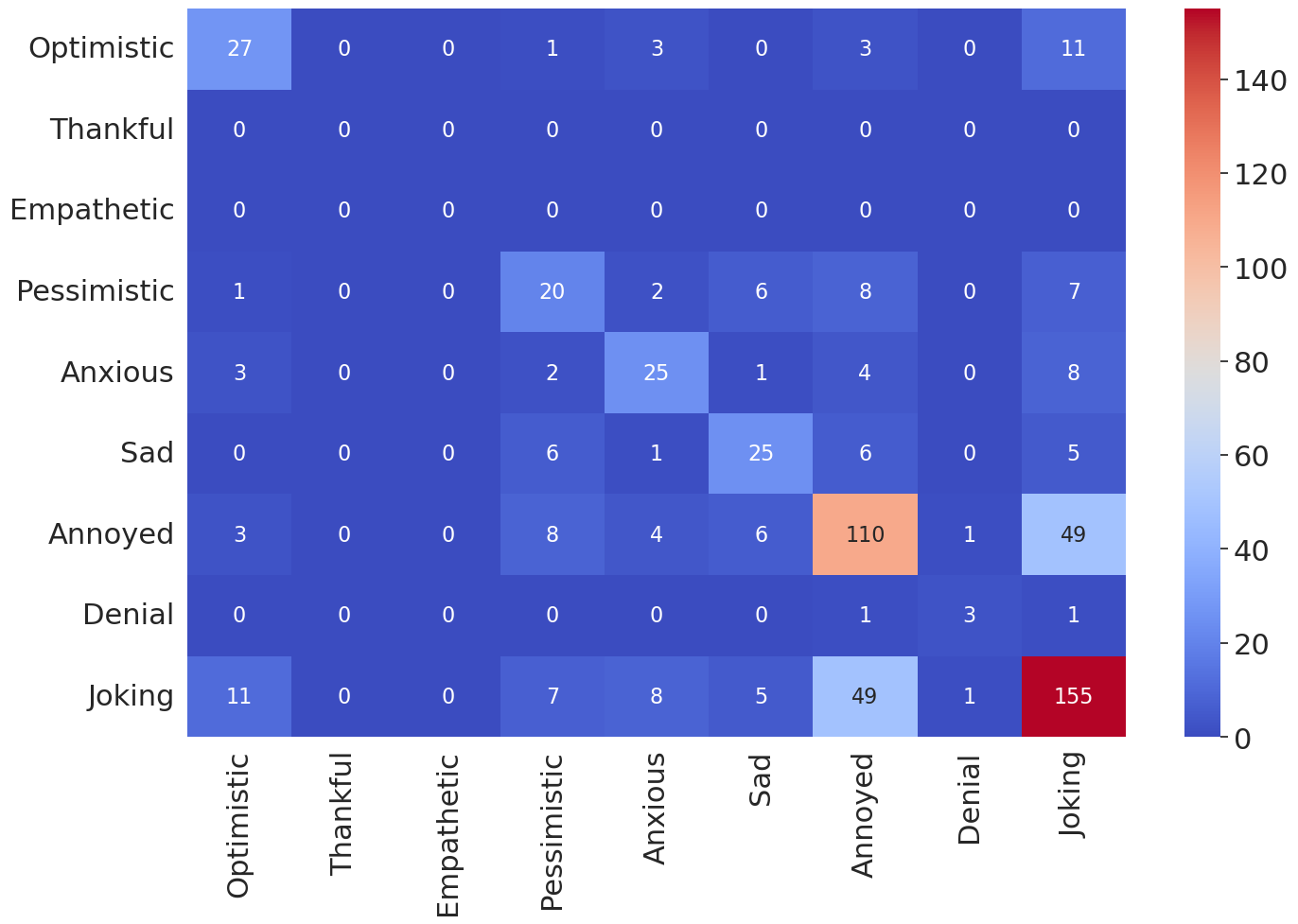}
        \caption{Julia}
        \label{fig:heatmap3}
    \end{subfigure}
    \caption{Heat map of different translations.}
    \label{fig:heatmap_n}
\end{figure}

Additionally, we show heat map plots depicting sentiment counts about other sentiments (Figure \boxedref{fig:heatmap_n}) and observe that  \textit{Annoyed} and \textit{Joking} are high-frequency sentiments in all the three translations. Moreover, the three heat maps  indicate that [\textit{annoyed}, \textit{joking}] have the highest correlation compared to other pairs. In Figures \boxedref{fig:heatmap1} and \boxedref{fig:heatmap2}, we notice that [\textit{sad}, \textit{joking}] are the second highest correlated and in contrast, [\textit{optimistic}, \textit{joking}] is the second highest correlated in Figure \boxedref{fig:heatmap3}.

\begin{table}[htbp!]
    \centering
    \small
    \begin{tabular}{|>{\centering\arraybackslash}m{1.8cm}|>{\centering\arraybackslash}m{1.8cm}|>{\centering\arraybackslash}m{1.8cm}|>{\centering\arraybackslash}m{1.8cm}|>{\centering\arraybackslash}m{1.8cm}|}
        \hline
        \textbf{Chapters} & \textbf{GT-Xianyi and Gladys} & \textbf{GT-Julia} & \textbf{Xianyi and Gladys-Julia} \\
        \hline
        Chapter 1 & 0.583 & 0.569 & 0.722 \\
        \hline
        Chapter 2 & 0.592 & 0.571 & 0.722 \\
        \hline
        Chapter 3 & 0.627 & 0.589 & 0.691 \\
        \hline
        Chapter 4 & 0.587 & 0.555 & 0.662 \\
        \hline
        Chapter 5 & 0.585 & 0.56 & 0.67 \\
        \hline
        Chapter 6 & 0.551 & 0.525 & 0.623 \\
        \hline
        Chapter 7 & 0.573 & 0.534 & 0.6 \\
        \hline
        Chapter 8 & 0.555 & 0.519 & 0.599 \\
        \hline
        Chapter 9 & 0.536 & 0.507 & 0.587 \\
        \hline
        \textbf{Average} & \textbf{0.577} & \textbf{0.548} & \textbf{0.653} \\
        \hline
    \end{tabular}
    \caption{Sentiment analysis of selected pairs of translations by Google Translate (GT) with Jaccard similarity score of the predicted sentiments for all nine chapters. We provide the average Jaccard score at the bottom.}
    \label{tab:jaccard}
\end{table}


Finally, we measure the diversity and similarity of predicted sentiments, verse-by-verse for three translations. Table \boxedref{tab:jaccard} shows the sentiment analysis of selected pairs of translations with Jaccard similarity score for all nine chapters. The average highest score across the entire nine chapters is 0.653 from the pairs of Xianyi and Gladys' version and Julia's version. It further indicates that the human experts have the highest overlap of sentiments for the entire text. The selected pairs by Google Translate are not as high as those of the expert translations, thus indicating that Google Translate has not been as efficient as human experts in translating \textit{The Real Story of Ah-Q}.


\subsection{Semantic analysis}

Next, we present a framework for semantic analysis by comparing three translations. Firstly, we provide a verse-by-verse cosine similarity analysis for each chapter by utilising an MPNet-based model and encoding all verses. In Table \boxedref{tab:cosine}, we present a mean score with standard deviation (in a basket) where we observe that chapter nine is the most semantically similar, whereas chapter six is the least semantically similar across the three translations. Furthermore, we find that the score of the selected pairs for Xianyi and Gladys' with Google Translate is higher than the selected pairs for Julia's version with Google Translate. Additionally, we observe that in Table \boxedref{tab:jaccard}, the average score of selected pairs for Xianyi and Gladys' version with Google Translate is higher than the selected pair for Julia's version with Google Translate. Finally, we compare the average score of selected pairs between the expert translations  as a benchmark to evaluate Google Translate. We find that the average scores of selected pairs with Google Translate are lower than the benchmark. Thus,  this may indicate that Google Translate is not as accurate as the human experts, even though their selected pairs' scores are close to the scores obtained by human experts.   

\begin{table}[htbp!]
\small
    \centering
    \begin{tabular}{|>{\centering\arraybackslash}m{1.8cm}|>{\centering\arraybackslash}m{1.8cm}|>{\centering\arraybackslash}m{1.8cm}|>{\centering\arraybackslash}m{1.8cm}|>{\centering\arraybackslash}m{1.8cm}|}
        \hline
        \textbf{Chapters} & \textbf{GT-Xianyi and Gladys} & \textbf{GT-Julia} & \textbf{Xianyi and Gladys-Julia} \\
        \hline
        Chapter 1 & 0.637(0.135) & 0.622(0.159) & 0.770(0.128) \\
        \hline
        Chapter 2 & 0.645(0.148) & 0.591(0.139) & 0.764(0.124) \\
        \hline
        Chapter 3 & 0.633(0.139) & 0.578(0.159) & 0.725(0.124) \\
        \hline
        Chapter 4 & 0.711(0.139) & 0.645(0.155) & 0.724(0.163) \\
        \hline
        Chapter 5 & 0.681(0.117) & 0.634(0.161) & 0.764(0.140) \\
        \hline
        Chapter 6 & 0.698(0.106) & 0.540(0.192) & 0.587(0.215) \\
        \hline
        Chapter 7 & 0.680(0.174) & 0.617(0.180) & 0.691(0.161) \\
        \hline
        Chapter 8 & 0.700(0.144) & 0.619(0.166) & 0.728(0.133) \\
        \hline
        Chapter 9 & 0.736(0.128) & 0.681(0.130) & 0.796(0.117) \\
        \hline
        \textbf{Average} & \textbf{0.577} & \textbf{0.548} & \textbf{0.653} \\
        \hline
    \end{tabular}
    \caption{Semantic analysis using cosine similarity for comparing selected pairs for three translations. We provide the mean score  with standard deviation(in brackets) for all the nine chapters and average cosine scores at the bottom.}
    \label{tab:cosine}
\end{table}

\begin{table*}[ht]
    \centering
    \small
    \footnotesize
    \begin{tabular}{|c|c|p{3cm}|p{3cm}|p{3cm}|c|c|c|}
        \hline
        Chapter & Verse & Hsien & GT & Julia & GT-Hsien & GT-Julia & Hsien-Julia \\
        \hline
        9 & 43 & December 1921 & December 1921 December. & December 1921 & 0.953 & 0.953 & 1.00 \\
        \hline
        7 & 38 & This came as a complete surprise to Ah Q, who could not help being taken aback. When the old nun saw that he had lost his aggressiveness, she quickly shut the gate, so that when Ah Q pushed it again he could not budge it, and when he knocked again there was no answer. & Ah Q was surprisingly unexpected, and he couldn't help it; the old nun saw that he lost his anger, and then closed the door quickly. When A Q pushed again, he couldn't open it, and when he hit it again, he didn't answer. & Ah-Q froze, stupefied by the unexpectedness of it all. Seeing the wind had been taken out of his sails, the elderly nun shut the gate as quick as she could. When Ah-Q recovered himself enough to give it a shove, it refused to budge; when he tried knocking again, there was no answer. & 0.908 & 0.884 & 0.931 \\
        \hline
        4 & 42 & Ah Q naturally agreed to everything, but unfortunately he had no ready money. Luckily it was already spring, so it was possible to do without his padded quilt which he pawned for two thousand cash to comply with the terms stipulated. After kowtowing with bare back he still had a few cash left, but instead of using these to redeem his felt hat from the bailiff, he spent them all on drink.  Actually, the Chao family burned neither the incense nor the candles, because these could be used when the mistress worshipped Buddha and were put aside for that purpose. Most of the ragged shirt was made into diapers for the baby which was born to the young mistress in the eighth moon, while the tattered remainder was used by Amah Wu to make shoe soles. & Ah Q naturally agreed, but unfortunately there was no money. Fortunately, in the spring, the quilt can be useless, and it has 2,000 money andfulfils the treaty. After shirtlessness, there were a few more articles left, and he no longer redeemed a felt hat, and he drank all. But the Zhao family did not burn incense candle, because the wife could use it when worshiping the Buddha, and kept it. The broken cloth shirt was mostly the lining of the children born in August in August, and thehalf-tattered sole was made. & Regrettably, Ah-Q lacked the funds to make good his indemnity. But as, by happy coincidence, it was spring, he was able to do without his cotton quilt, which he pawned for two thousand coppers, enabling him to fulfil the demands of the peace treaty. After kowtowing, bare-chested, he found himself with a few coppers left over, which he chose to blow on wine rather than redeem his felt hat. The Zhaos didn’t burn the candles and incense right away, preferring to keep them for when the mistress of the house next paid her respects to the Buddha. Most of his tattered old shirt was recycled into nappies for the baby that was born to the younger mistress in the eighth month; any off-cuts were used by Mrs Wu for the soles of her shoes. & 0.876 & 0.886 & 0.898 \\
        \hline
    \end{tabular}
    \caption{Semantically most similar verses using the cosine similarity (score) using selected translations for comparison (Hsien(Xianyi), Julia) vs Google Translate (GT).}
    \label{tab:similar-verses}
\end{table*}

\begin{table*}[t]
    \centering
    \begin{tabular}{|c|c|p{3cm}|p{3cm}|p{3cm}|c|c|c|}
        \hline
        Chapter & Verse & Hsien & GT & Julia & GT-Hsien & GT-Julia & Hsien-Julia \\
        \hline
        7 & 8 & Tra la, tra la! & Well!!  & Tum-ti-tum, clang clang-clang! & 0.139 & 0.154 & 0.554 \\
        \hline
        9 & 36 & So Ah Q took another look at the shouting crowd. & Ah Q, then look at those who drink. & Ah-Q looked back at the cheering crowd. & 0.268 & 0.168 & 0.715 \\
        \hline
        8 & 19 & "Get out!" said Mr. Foreigner, lifting the "mourner's stick." & "Go out!" Mr. Yang rose to cry. & "Get lost!" Mr Foreigner began waving his stick about. & 0.351 & 0.180 & 0.652 \\
        \hline
    \end{tabular}
    \caption{Semantically least similar verses using the cosine similarity (score) using selected translations for comparison (Hsien (Xianyi), Julia) vs Google Translate (GT).}
    \label{tab:least-similar-verses}
\end{table*}

Additionally, we present some semantically similar verses by using the cosine similarity score in Table \boxedref{tab:similar-verses}. In Chapter 9 - Verse 43, it is interesting to note that verses from Xianyi(and Gladys) and Julia are exactly the same. The choice of words from Google Translate is also the same as the two human translations, whereas there is an extra 'December' in the verse. In Chapter 7 - Verse 38 and Chapter 4 -Verse 42, all three translations have utilized some of the same or similar words such as 'nun', 'quickly', 'naturally', 'unfortunately', 'regrettably'. Thus, all selected pairs have obtained high scores (over 0.85 cosine similarity). On the other hand, we present some of semantically least similar verses by using the cosine similarity score in Table \boxedref{tab:least-similar-verses}. In Chapter 7 - Verse 8, we observe that the choice of mimetic  words for all three translations differs.  Moreover, for Chapter 9 - Verse 8 and Chapter 8 - 19, Google Translate and Julia Lovell conveys different meanings, and is therefore assigned a low similarity score.

\begin{table*}[t]
    \centering
    \begin{tabular}{|c|c|p{3cm}|p{3cm}|p{3cm}|c|c|c|}
        \hline
        Chapter & Verse & Hsien & GT & Julia & GT-Hsien & GT-Julia & Hsien-Julia \\
        \hline
        2 & 1 & In addition to the uncertainty regarding Ah Q's surname, personal name, and place of origin, there is even some uncertainty regarding his "background." This is because the people of Weichuang only made use of his services or treated him as a laughing-stock, without ever paying the slightest attention to his "background." Ah Q himself remained silent on this subject, except that when quarrelling with someone he might glance at him and say,"We used to be much better off than you! Who do you think you are anyway?" & A Q is not unique to the name of his name, and even his previous "line" is also slim. Because the people of Wei Zhuang are in Ah Q, as long as he helps, he just joked with him, and never pays attention to his "line". And A Q himself did not say that when he was unique to others, he or stared at his eyes: "We used to be much wider than you! What are you!"  & It was not only Ah-Q’s name and place of origin that were shrouded in mystery – but also the details of his early life. Because the good people of Weizhuang called upon him only to help out with odd jobs, or to serve as the butt of jokes, no one ever paid much attention to such niceties. Neither was Ah-Q himself particularly forthcoming on the subject, except when he got into arguments, viz.: ‘My ancestors were much richer than yours! Scum!’ & 0.715 & 0.678 & 0.883 \\
        \hline
        2 & 7 & Ah Q would rise to the bait as usual, and glare furiously. & Ah Q became angry as usual, and he watched angrily. & Cue the Angry Glare. & 0.733 & 0.351 & 0.453 \\
        \hline
        2 & 16 & "The Passage—one hundred—one hundred and fifty." & "One hundred in the hall -one hundred and fifty!" & "One hundred on the Passage – one hundred and fifty!" & 0.668 & 0.742 & 0.889 \\
        \hline
        2 & 21 & So white and glittering a pile of silver! It had all been his... but now it had disappeared. Even to consider it tantamount to being robbed by his son did not comfort him. To consider himself as an insect did not comfort him either. This time he really tasted something of the bitterness of defeat. & A pile of white money! And it was him -now it's gone! It is said that he was taken away by his son, and he was always unhappy; he said that he was a worm.Only to feel the pain of failure. & That shiny pile of silver dollars! Once it had been all his – but where was it now? He tried telling himself his son had stolen it; his discontent continued to simmer. He told himself he was a slug – still no peace of mind. Now, only now, did he feel the bitterness of defeat. & 0.588 & 0.560 & 0.722 \\
        \hline
    \end{tabular}
    \caption{Semantic similarity score from Chapter 2 with cosine similarity to compare arbitrarily selected verses by selected human experts translations to Google Translate}
    \label{tab:selected-chapter-verses}
\end{table*}

Finally, we investigate arbitrarily selected verses from one chosen chapter using cosine similarity for all three translations. In this case, apart from Chapter 1 (Introduction), we observe Chapter 2 as it includes fewer verses an is easier to analyze. In  Table \boxedref{tab:selected-chapter-verses}, we arbitrarily selected some of the verses to compare their cosine similarity for all three translations in Chapter 1. 

\section{Evaluation by human expert}

Next, we investigate the comparison of Google Translate's translations and human experts' translations. We show a more detailed analysis to gain a deeper understanding of the advantages worth mentioning as well as disadvantages worth noting for Google Translate. We evaluate the translations by human experts who coauthor this paper, Rodney Beard and Xuechun Wang. 

\subsection{Semantically most similar verses}

In Table \boxedref{tab:similar-verses}, we present the most similar semantic verses across all nine chapters by using cosine similarity with original texts \textit{The True Story of Ah-Q} in Chinese characters (Figure \boxedref{tab:chinese-most-verses}). 

In Chapter 9: Verse 43, the three pairs present high scores in cosine similarity for Hsien -Julia since the choices of words are the same. In the Google Translate version, it also presents 'December 1921' in the sentence; however, an extra 'December' follows it. The sentence \textit{一九二一年十二月}, exactly presents 'December 1921', thus both human expert translations represent an accurate meaning. However, Google Translate has an extra \textit{十二月} 'December' and hence provide a poor sentence structure.

In Chapter 7: Verse 38,  Google Translate(GT)-Hsien is more semantically similar to GT-Julia as shown in  Table \boxedref{tab:chinese-most-verses}. The words '锐气' describe ergetic determination, ambition, or vibrant drive. In the three translations for  the phrase \textit{他失了锐气}, Xianyi (and Gladys)'s  describes it as \textit{'he had lost his aggressiveness'}, whereas Julia describes it as \textit{'the wind had been taken out of his sails'}. Google Translate    describes it as \textit{'he lost his anger'} since anger is more frequently used  to describe someone's annoyance, rather than a vibrant drive. Thus, Google Translate conveys a misunderstanding in the meaning of this phrase. On the other hand, although the two expert translations do not directly translate the term '锐气', they convey its intended meaning contextually.

In Chapter 4: Verse 42, the cosine similarity score for the three translations is nearly equal to each other, and GT-Julia is slightly higher than GT-Hsien. In this case, we notice that Google Translate misunderstand the phrase '居然还剩几文'. '居然还剩几文' could be described as there is still a few coppers left. The character '文' does not represent an article, but rather money. The two  expert have accurately translated '文' as cash (from Xianyi and Gladys) and coppers (from Julia) respectively. However, Google Translate has translated '文' as 'article' which is wrongly translated and lacks contextual significance.

\begin{table*}[htbp]
    \centering
    \footnotesize
    \begin{tabular}{|c|c|p{12cm}|}
        \hline
        Chapter & Verse & Original texts\\
        \hline
        9 & 43 & 一九二一年十二月。 \\
        \hline
        7 & 38 &  阿Q很出意外，不由的一错愕;老尼姑见他失了锐气，便飞速的关了 门，阿Q再推时，牢不可开，再打时，没有回答了。 \\
        \hline
        4 & 42 & 阿Q自然都答应了，可惜没有钱。幸而已经春天，棉被可以无用，便质了二千大钱，履行条约。赤膊磕头之后，居然还剩几文，他也不再赎毡帽，统统喝了酒了。但赵家也并不烧香点烛，因为太太拜佛的时候可以 用，留着了。那破布衫是大半做了少奶奶八月间生下来的孩子的衬尿布， 那小半破烂的便都做了吴妈的鞋底。\\
        \hline
    \end{tabular}
    \caption{Semantically most similar verses across translations in Table \boxedref{tab:similar-verses} showing original texts \textit{The True Story of Ah-Q} in Chinese characters.}
    \label{tab:chinese-most-verses}
\end{table*}

\begin{table*}[htbp]
    \centering
    \begin{tabular}{|c|c|p{12cm}|}
        \hline
        Chapter & Verse & Original texts\\
        \hline
        7 & 8 & 得得，锵锵! \\
        \hline
        9 & 36 &  阿Q于是再看那些喝采的人们。 \\
        \midrule
        8 & 19 & “滚出去!”洋先生扬起哭丧棒来了。\\
        \hline
    \end{tabular}
    \caption{Semantically least similar verses across translations in Table \boxedref{tab:least-similar-verses} showing original texts \textit{The True Story of Ah-Q} in Chinese characters.}
    \label{tab:chinese-least-verses}
\end{table*}

\subsection{Semantically least similar verses}

In Table \boxedref{tab:least-similar-verses}, we present the least similar verses across all nine chapters by using cosine similarity with original texts \textit{The True Story of Ah-Q} in Chinese characters (Figure \boxedref{tab:chinese-least-verses}). We find that in some cases, the Google Translate version conveys a misunderstanding or lack of contextual significance in sentences. In Chapter 7: Verse 8 of Table \boxedref{tab:least-similar-verses}, GT-Hsien has the lower cosine similarity   than GT-Julia. According to Figure \boxedref{tab:chinese-least-verses}, '得得，锵锵' as an onomatopoeia \cite{assaneo2011anatomy} which primarily mimics the sounds. The GT version does not suggest a sound with no logical sense. On the other hand, Hsien and Julia versions  attempt to create mimetic words \cite{gu2005mimetic} to convey a logical meaning.

In Chapter 9: Verse 36 of Table \boxedref{tab:least-similar-verses}, both GT-Hsien and GT-Julia have low cosine similarity with GT-Julia showing semantically least similar verses. However,  Hsien-Julia represents a relatively high cosine similarity than those pairs with Google Translate. Some of the words are wrongly translated by Google Translate, including the verb '喝采' which could be described as 'cheer up'. The single character '喝' as a verb means 'drink', e.g. I drink water) Google Translate has omitted the character '采'，as it only conveys the meaning of the single character '喝'; however, the verbs '喝' and '喝采' have entirely different meanings. Thus, it led to the wrong translation of the entire sentence and bereft of logic and contextual significance.  

In Chapter 8 - Verse 19 of Table \boxedref{tab:least-similar-verses}, Google Translate version lacks logic and contextual significance, e.g., '洋先生' from the original texts (Figure \boxedref{tab:chinese-least-verses}) describes a man who comes from a foreign country, thus we could call him 'Mr Foreigner' because one of the meanings of '洋' is 'foreign'. However, Google Translate lacks contextual significance and only conveys literal meaning as someone's surname 'Yang'. Moreover, the phrase '扬起哭丧棒来了' has been omitted by Google Translate and '哭丧' has been omitted by Julia. Only translations by Xianyi and Gladys are contextually significant for this sentence.

\subsection{Arbitrarily selected verses}

Chapter two is one of the important chapters in \textit{The True Story of Ah Q} since it describes the characteristics of Ah Q and the beginning of the story. We arbitrarily select four verses to compare their similarity scores (from Table \boxedref{tab:selected-chapter-verses}). In general, we find that the Hsien-Julia similarity score is higher than those pairs with Google Translate.

Chapter 2: Verse 1, GT-Hsien is semantically most similar, as evidenced by its cosine similarity score and both Hsien and Julia represent a contextual and logical significance. However, the sentence\textit{'A Q is not unique to the name of his name, and even his previous ”line” is also slim.'} by  Google Translate is unclear in its meaning and is significantly different compared with the two human expert translations. Thus, it is still bereft of contextual and logical significance.  The Chapter 2: Verse 7, GT-Hsien is semantically most similar and the cosine similarity score by GT-Julia and Hsien-Julia is much lower than GT-Hsien. Thus, we can infer that Julia's translation of this verse exhibits significant semantic differences compared to the GT version and the Hsien version. The sentence \textit{'Cue the angry glare'} by Julia has omitted its translations. Conversely, although the choices of words by GT and Hsien versions are different, they both maintain contextual and logical significance. 

Chapter 2: Verse 16, GT-Julia is semantically most similar, as evidenced by its cosine similarity score. The \textit{passage} or \textit{hall} describes betting on the passage or hall (it is a phrase used when placing a bet), and we should use the preposition \textit{'on'} instead of \textit{'in'} and if changed, the meaning of the sentence would be different. Hence, Google Translate is unable to understand the context of words and lack of contextual significance. Chapter 2: Verse 21, the cosine similarity scores of GT-Hsien and GT-Julia are nearly equal, with GT-Julia showing minimum semantic similarity. The sentence \textit{'And it was him - now it’s gone! It is said that he was taken away by his son, and he was always unhappy'} is inaccurately translated by Google Translate. The original texts describe that although the money might have been stolen by his son, he still felt uncomfortable. The Google Translate version maintains logical sense but loses contextual significance.

\section{Discussion}

We can observe from the semantic analysis that Google Translate often mistranslated verses, especially when the verses are contextually related or have strong logical connections. Moreover, Google Translate sometimes struggles to accurately segment sentences; e.g. '阿Q于是再看那些喝采的人们。' from Table \boxedref{tab:chinese-least-verses}, Google Translate incorrectly segmented the verb '喝采' into '喝' and ignored '采'. Additionally, in some cases, Google Translate does not work on certain words, causing some chapters in the translated version with original Chinese characters. In Chapter 2-Verse 11, '倘他姓赵，则据现在好称郡望的老例，可以照《郡名百家姓》上的注解，说是“陇西天水人也”，但可惜这姓是不甚可靠的，因此籍贯, there is also some uncertainty. Although he lives more, he often stays elsewhere.', Google Translate was unable to identify this sentence in Chinese, so it has omitted translating it into English.

 Google Translate has an inability to accurately translate certain phrases or sentences, we infer that this is due to its lack of capability to consider context for translation. Additionally, Google Translate is not comprehensive enough to translate some vernacular Chinese. Particularly with the texts used in our study, which are formed in a narrative format, each chapter is interrelated. This may raise a potential challenge that if the choices of words, or the contents of verses or each chapter are translated individually, it can lead to inaccuracies.  Zhao Taiye (赵太爷) is one of the characters throughout the whole story, but his name is not translated consistently the same in Google Translate, and we provide some examples  below:

\begin{itemize}
    \item Original texts: Chapter 6 - Verse 1: 然而未庄老例，只有\textbf{赵太爷} 钱太爷和秀才大爷上城才算一件事。 
    \item Google Translate: Chapter 6 - Verse 1: However, in the old case of Wei Zhuang, only the \textbf{grandfather Zhao} and uncle Xiucai Shangcheng were considered one thing.
\end{itemize}

\begin{itemize}
    \item Original texts: Chapter 7 - Verse 25:  “这时未庄的一伙鸟男女才好笑哩，跪下叫道，‘阿Q，饶命!’谁听他! 第一个该死的是小D和赵太爷，还有秀才，还有假洋鬼子，......留几条 么?王胡本来还可留，但也不要了。......
    \item Google Translate: Chapter 7 - Verse 25: "At this time, a group of birds and women in Weizhuang were funny, kneeling down," Ah Q, Life! 'Who listened to him! The first one damn is Xiao D and \textbf{Master Zhao}, there are Xiucai, and fake foreign devils, ... Is there a few? Wang Hu can still stay, but don't.
\end{itemize}

If we only focus on one verse, it is considerably correct to be translated as Grandfather Zhao or Master Zhao. However, it can cause confusion for readers, if the two translated names are both represented in translations and are exactly for one character.

Moreover, \textit{The True Story of Ah Q} is a novella written in vernacular Chinese. In the novella, apart from single Chinese characters, it also incorporates four-character idioms (成语 chengyu) or Chinese allusions. For example, '塞翁失马安知非福'（sai weng shi ma an zhi fei fu）from Chapter 2, the original text is '但真所谓“塞翁失马安知非福”罢，阿Q不幸而赢了一回，他倒几乎失败 了。'。 '塞翁失马安知非福'（'sai weng shi ma an zhi fei fu'）originally comes from Huainan Zi in the Chapter of Ren Jian Xun. The story behind this allusion conveys an old man 'Sai Weng,' who lost a horse. While others felt sorry for him, he thought that this bad thing could turn out to be a good thing. Then, his horse not only returned but also brought back a fine horse. Everyone thought this was a good thing, but Sai Weng believed that a good thing could also turn into a bad thing. Subsequently, his son broke his leg while riding the horse. Though people came to offer their condolences, Sai Weng felt that breaking a leg was a minor misfortune that saved his son's life. Later, when the Xiongnu invaded and the state began recruiting soldiers, most of those who went to war died on the battlefield. However, Sai Weng's son, because of his broken leg, did not enlist in the army and thus survived. The story implies that misfortune may lead to good outcomes or good fortune may conceal adverse consequences. Although Ah Q couldn't continue gambling due to lack of money, it actually prevented him from losing more money, which turned out to be a good thing for him. This circumstance corresponds to the allusion '塞翁失马安知非福'（'sai weng shi ma an zhi fei fu'). This verse (Chapter 2 - Verse 19)  has been translated by Google Translate as 'But the so -called "Sai Weng lost Ma An Zhifu"', Ah Q won the unfortunate time, and he almost failed.' This illustrates that Google Translate did not convey the intended meaning; it merely translated the allusion from Chinese characters into pinyin. On the other hand, the verse has been translated by Xianyi and Gladys as 'However, the truth of the proverb "misfortune may be a blessing in disguise" ' was shown when Ah Q was unfortunate enough to win and almost suffered defeat in the end.'. This translation conveys the cultural significance of the allusion to readers.

Our study has three major limitations. Firstly, we only use one type of text (novel) in our study, which would considerably reduce the authenticity of the comparison between Google Translate and human experts. For further improvement, we could use various types of texts such as newspapers, official reports, and academic papers to reduce bias. Secondly, we fine-tuned BERT-based uncased model refined using the SenWave dataset \cite{yang2020senwave}. However, SenWave dataset contahuman-labelledlled tweets about COVID-19 that may be inappropriate for capturing sentiments in novels. The novel used in our study is a century old, and the vocabulary of the transitions is also widely different when compared to the expression used in social media such as Twitter (X). Other forms of labelled sentiment analysis datasets could be considered, and more effort and time would be needed for quantitative analysis to improve our study. Finally, based on our cumulative sentiment analysis \boxedref{fig:combo_combo}, \textit{joking} is over-reported and in comparison to a related study of sentiment analysis of newspaper articles during COVID-19  \cite{chandra2024large}, \textit{joking} did not count as over-reported emotion. Newspaper articles have a different style when compared to expressions used in novels, with different situations displaying various sentiments from different characters as the plot unfolds. Our study would present different sentiments for a novel with romance and comedy since the drama category of novels captures difficult situations in life  and generally captures more negative emotions. \cite{trosborg1997text} \cite{lee2001genres} In \textit{The True Story of Ah Q}, the texts convey a sense of irony rather than joking. However, since SenWave dataset does not contain irony emotion, it may have categorised irony as joking. This is also a major limitation due to the lack of emotions in our utilised dataset. In future works, a diverse sentiment-labelled dataset can be manually created through crowdsourcing.          




\section{Conclusions}

We presented a framework for evaluating translations between Google Translate and human experts from Chinese Mandarin to English via sentiment and semantic analysis. In the three translations, we first presented bigrams and trigram analysis   and then  conducted a chapter-wise sentiment and semantic analysis to investigate extracted sentiments across three translations. 


We found that \textit{thankful} and \textit{empathetic} were not expressed in any of the chapters and Google Translate led sentiments that expressed \textit{sad} and \textit{joking}. Moreover, Julia Lovell translation led\textit{Optimistic}, \textit{Pessimistic} and \textit{Anxious} sentiments while   Xianyi Yang and Gladys T Yang led the   \textit{Annoyed} sentiment. The differences in the sentiments by the different translators show biases of transitions and also unique style of vocabulary and words used to capture the various situations. The vocabulary also changes for different decades, both in the written and oral transmission of languages. Overall, based on the bigrams and trigrams plot, we found that although the choices of words by three translations are different, the differences in sentiment among the three translations are not vastly different. Furthermore, in terms of semantic analysis, we found that Chapter 9 was the most semantically similar, whereas Chapter 6 is the least semantically  in the  three translations. Moreover, we found that both in sentiment and semantic analysis, the similarity between Google Translate and  human expert is generally lower than the similarity between two human experts translations. 

The inaccuracy of Google Translate could be due to the lack of logical and contextual significance. Novella as a type of text depends on contextual connections for accurate translation. We found that  Google Translate has the inability to maintain the contextual connections for translating from Mandarin to English. Furthermore,  Google Translate cannot translate Chinese allusions accurately.


In the future studies, we can either evaluate Google Translate in other languages, and current framework can be extended to different types of texts such as newspapers, or academic papers in Chinese to English translations.  



\section*{Code and Data Availability}
GitHub repository featuring open source code and data \footnote{\url{https://github.com/sydney-machine-learning/translationanalysis-Mandarin}}.

 \bibliographystyle{elsarticle-num} 
 \bibliography{cas-refs}





\end{CJK*}
\end{document}